\documentclass[10pt,twocolumn,letterpaper]{article}

\usepackage[pagenumbers]{cvpr} % To force page numbers, e.g. for an arXiv version

\usepackage{fancyhdr}
\fancypagestyle{firstpage}{
  \fancyhf{}
  
  \fancyhead[C]{\small This paper has been accepted for publication at the \\ IEEE/CVF Conference on Computer Vision and Pattern Recognition (CVPR), Denver, 2026.}
}

\usepackage{microtype}

\definecolor{cvprblue}{rgb}{0.21,0.49,0.74}
\usepackage[pagebackref,breaklinks,colorlinks,allcolors=cvprblue]{hyperref}
\usepackage{algorithmic}
\usepackage{multirow}
\usepackage{amsmath}
\DeclareMathOperator*{\argmax}{arg\,max}
\def\paperID{} % *** Enter the Paper ID here
\def\confName{CVPR}
\def\confYear{2026}

\title{The Missing GAP: From Solving Square Jigsaw Puzzles to Handling Real World Archaeological Fragments} %% We need to figure out how to incorporate the flow matching concept here. Unfortunately I already like this title.

\author{
    Ofir Itzhak Shahar \quad Gur Elkin \quad Ohad Ben-Shahar \\
    Stein Faculty of Computer and Information Science \\
    Ben-Gurion University of the Negev, Israel \\
    {\tt\small \{shofir, gurshal\}@post.bgu.ac.il, ben-shahar@cs.bgu.ac.il}
}

\begin{document}
\maketitle
\thispagestyle{firstpage}
\begin{abstract}
Jigsaw puzzle solving has been an increasingly popular task in the computer vision research community. Recent works have utilized cutting-edge architectures and computational approaches to reassemble groups of pieces into a coherent image, while achieving increasingly good results on well established datasets. However, most of these approaches share a common, restricting setting: operating solely on strictly square puzzle pieces. In this work, we introduce GAP, a set of novel jigsaw puzzles datasets containing synthetic, heavily eroded pieces of unrestricted shapes, generated by a learned distribution of real-world archaeological fragments. We also introduce PuzzleFlow, a novel ViT and Flow-Matching based framework for jigsaw puzzle solving, capable of handling complex puzzle pieces and demonstrating superior performance on GAP when compared to both classic and recent prominent works in this domain. 
\end{abstract}

\section{Introduction}
Solving jigsaw puzzles has been one of humanity's preferred casual hobbies for many centuries. Not very surprisingly, it was established as a computational task in the early 1960's \cite{freeman1964apictorial}, and has been an active research topic ever since. 

Throughout the decades, this computational task has evolved far beyond its origins as a recreational activity and have been utilized for numerous applications, both within and outside computer science. Within computer science, these include digital security~\cite{gao2010novel,ali2014development}, solving instances of other NP-hard problems~\cite{Zhao2020aJigsaw}, being utilized as an unsupervised learning objective for training deep neural networks~\cite{noroozi2016unsupervised, misra2020self, wei2019iterative} and even as an auxiliary task to improve model generalization~\cite{carlucci2019domain, chen2023jigsaw}. Beyond computer science, jigsaw puzzle solving has been applied to biology~\cite{gassner1996test,marande2007mitochondrial}, paleontology~\cite{warren2014puzzle}, forensics~\cite{ukovich2004shredded, xu2014solution}, and most prominently, archaeology and the reconstruction of broken artifacts. In fact, the latter is repeatedly presented as a main motivation for addressing puzzle solving computationally~\cite{willis2008computational,kleber2009scientific,sizikova2017wall, tsesmelis2024re, derech2021solving, shahar2025pairwise, safaei2025solving, islam2025reassemblenet, rika2019novel, da2002multiscale, harel2024pictorial, ohayon2025solving}. 
% However, whilst the puzzle-solving frameworks have continuously progressed, following the overall progression in computer vision research throughout the last decade, the general setting these frameworks aimed to solve remained relatively still. Most existing puzzle-solving approaches work on a somewhat simplistic instance of this problem, addressing solely pieces of regular square shapes, with very minimal regard to any element of erosion or potential gap between puzzle pieces, representing it, if at all, mostly as a fixed gap between pieces~\cite{song2023siamese}. 
And yet, while puzzle-solving frameworks have continuously progressed alongside advances in computer vision research, the problem settings they address have remained largely intact. Indeed, most existing approaches operate on a simplified instance of the problem, addressing exclusively square-shaped pieces with little to no consideration of erosion or variable spacing between fragments. When gaps are modeled at all, they are typically represented as fixed uniform spacing~\cite{song2023siamese}.
\begin{figure}[h]
\centering
\includegraphics[width=0.9\linewidth]{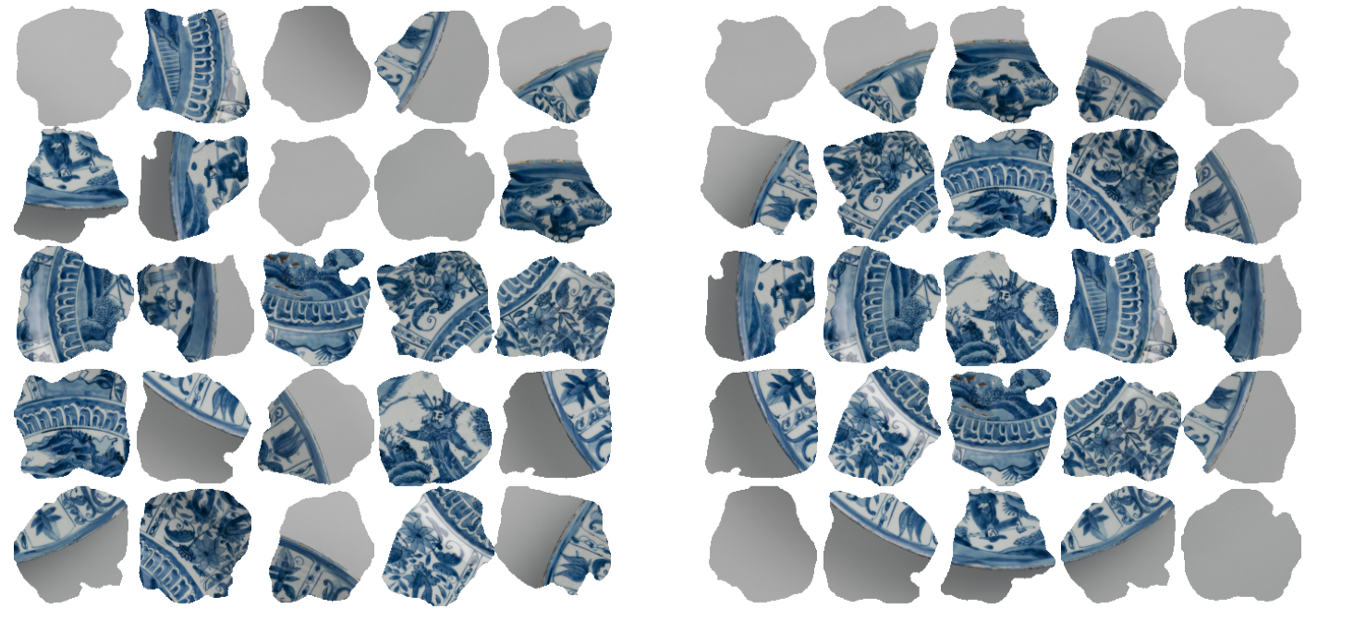}
\caption{\textbf{Archaeological Puzzle Reconstruction.} A puzzle from GAP-5 dataset (left) features irregularly-shaped, heavily eroded fragments generated from real archaeological artifact distributions. PuzzleFlow (right) successfully reconstructs these challenging puzzles by learning holistic visual relationships across entire fragment surfaces, rather than relying on boundary continuity. 
% The puzzle was made with an image of a Porcelain painted with cobalt blue under a transparent glaze Ming dynasty, Chongzhen period.
}
\label{fig:teaser}
\end{figure}
In this work, we address this fundamental limitation through two complementary contributions. First, we introduce \textbf{GAP} (Generated Archaeological-fragments Puzzles), a benchmark dataset featuring puzzles with heavily eroded, irregularly-shaped fragments that aim to capture the geometric complexity of real-world archaeological reconstruction. GAP puzzle pieces are generated via a Variational Autoencoder trained on authentic archaeological fragments from the RePAIR dataset~\cite{tsesmelis2024re}, producing synthetic fragments that preserve the statistical distribution of real artifact morphologies while enabling large-scale dataset creation. Second, we introduce \textbf{PuzzleFlow}, a novel framework that leverages Vision Transformers and discrete flow matching to solve jigsaw puzzles with arbitrary fragment geometries. Unlike prior approaches that often rely on matching content along fragment boundaries -- a strategy that fails when erosion eliminates the original edge information -- our architecture enables holistic relational reasoning across entire fragment surfaces, learning to identify global visual patterns, color distributions, structural coherence, and boundary characteristics that transcend local boundary features. As demonstrated in Fig.~\ref{fig:teaser}, PuzzleFlow successfully reconstructs complex archaeological-like puzzles where traditional boundary-matching methods fail. Together, these contributions bridge the gap between simplified academic puzzle settings and the challenging requirements of practical heritage reconstruction applications.

% Unlike prior approaches that rely on matching content along fragment boundaries—a strategy that fails when erosion has destroyed edge information—our architecture enables holistic relational reasoning across entire fragment surfaces through cross-piece attention

% a novel framework that leverages Vision Transformers with cross-piece attention mechanisms and discrete flow matching

% We address this gap through two contributions. We present GAP (Generated Archaeological Puzzles), benchmark datasets containing irregularly-shaped, heavily eroded fragments generated by a VAE trained on 958 real archaeological pieces, preserving authentic artifact complexity while enabling large-scale evaluation. We then introduce PuzzleFlow, a Vision Transformer-based framework employing cross-piece attention and discrete flow matching. Rather than relying on boundary continuity—which erosion destroys—PuzzleFlow learns holistic relationships by attending to global visual and geometric patterns distributed across entire fragment surfaces. This approach demonstrates superior performance on GAP compared to adapted baselines, marking a significant step toward practical archaeological reconstruction systems. As demonstrated in Fig.~\ref{fig:teaser}, PuzzleFlow successfully reconstructs complex archaeological-like puzzles where traditional boundary-matching methods fail. Together, these contributions bridge the gap between simplified academic puzzle settings and the challenging requirements of practical heritage reconstruction applications

\section{Related Work}
\label{sec:related_work}

% Decades after the initial formulation of puzzle-solving as a computational task \cite{freeman1964apictorial}, and furthermore after it was established as NP-complete~\cite{demaine2007jigsaw}, the problem has attracted sustained attention from the computer vision community. The field has witnessed a progression from classical optimization methods to sophisticated learning-based frameworks, with most contemporary research concentrating on square-piece configurations due to problem complexity. As we aim to utilize and expand this common setting, we focus on works which rely on this setting. For further elaboration on puzzle solving frameworks on different settings, we refer the reader to recent surveys~\cite{lu2025survey, tsesmelis2024re, shahar2025pairwise, markaki2023jigsaw}.

Since its early conception as a computational challenge~\cite{freeman1964apictorial}, and its later recognition as an NP-complete problem \cite{demaine2007jigsaw}, puzzle-solving has remained a compelling pursuit in computer vision research. Over the decades, the field has witnessed a progression from hand-crafted optimization schemes to powerful, data-driven learning frameworks, with most recent efforts centering around square-piece jigsaw puzzles. While here we aim to cover the extensive body of work addressing this variant,  more information about alternative puzzle shapes can be found in recent surveys~\cite{lu2025survey, tsesmelis2024re, shahar2025pairwise, markaki2023jigsaw}.

% It should be emphasized that given the complexity of this task, the vast majority of works have chosen to focus on limited instances of puzzles, containing solely pieces of identical square shapes. As we aim to utilize and expand this common setting, we focus on works which rely on this setting. For further elaboration on puzzle solving frameworks on different settings, we refer the reader to recent surveys~\cite{lu2025survey, tsesmelis2024re, shahar2025pairwise, markaki2023jigsaw}.

\begin{figure}[t]
\centering
\includegraphics[width=0.32\linewidth]{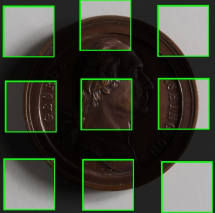}
\hfill
\includegraphics[width=0.32\linewidth]{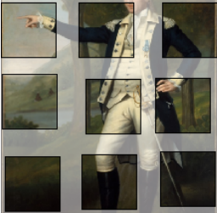}
\hfill
\includegraphics[width=0.32\linewidth]{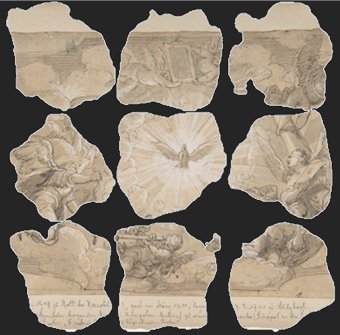}
\caption{Visual comparison of puzzle erosion patterns across datasets. \textbf{Left:} JPwLEG-3~\cite{song2023siamese} features square pieces with fixed 44px uniform gaps. \textbf{Center:} Deepzzle~\cite{paumard2020deepzzle} employs square pieces with random linear gap along edges. \textbf{Right:} Our GAP-3 dataset exhibits irregular fragment geometries with variable, non-linear gaps that are learned from real archaeological erosion patterns. The presented image is 'The Holy Ghost Surrounded by Angels' by Hans Georg Asam}
\label{fig:dataset_comparison}
\end{figure}

\begin{figure*}[h]
    \centering
    \includegraphics[width=0.9\textwidth]{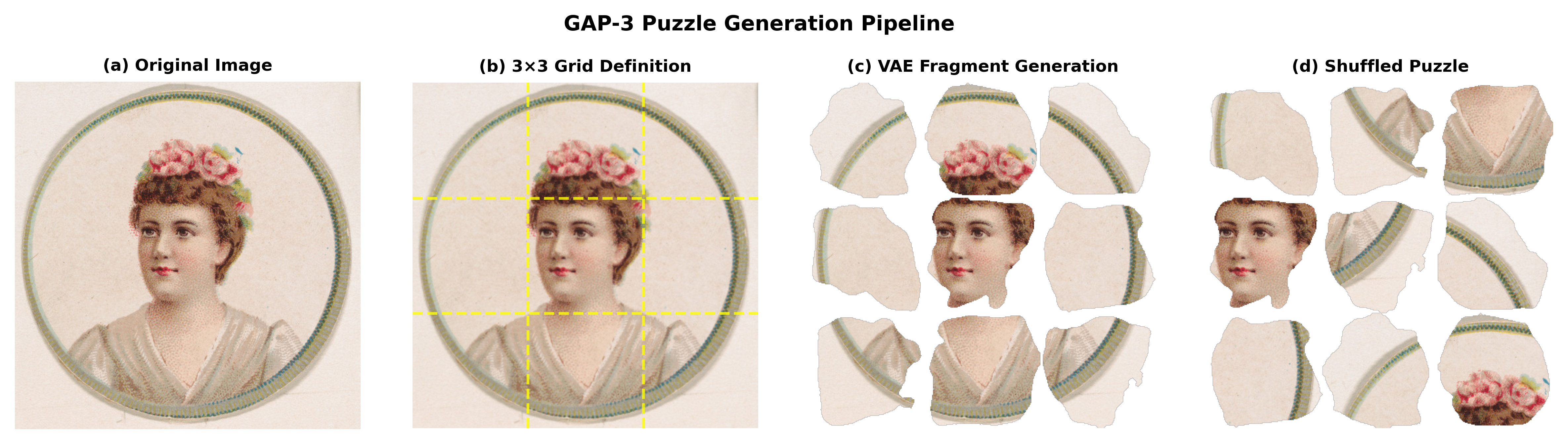}
    \includegraphics[width=\textwidth]{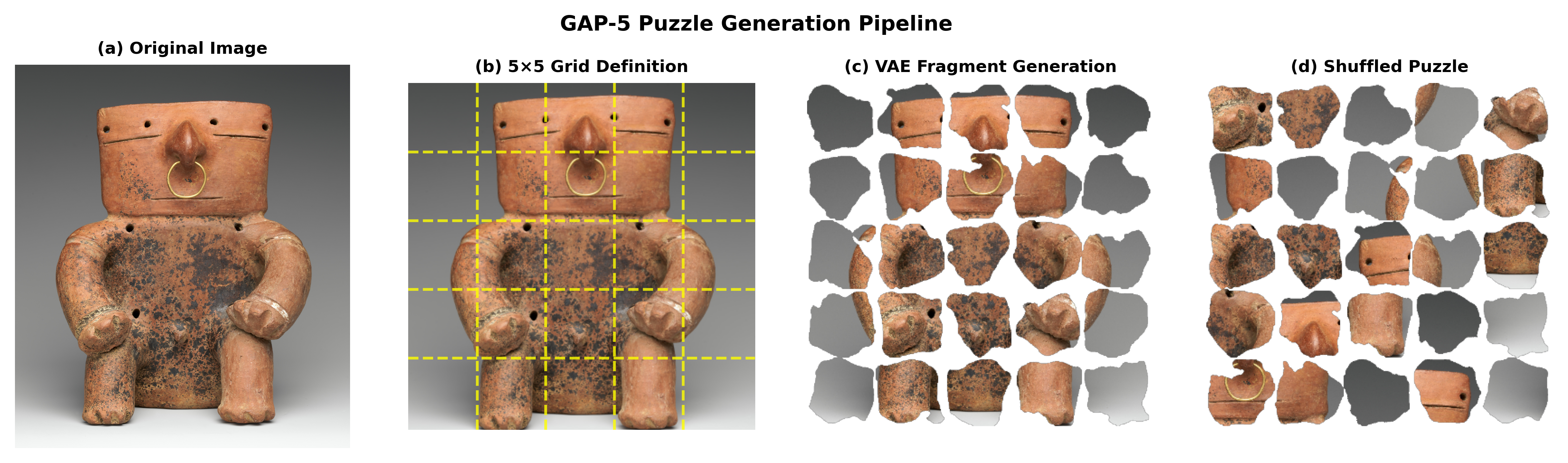}  
    \caption{\textbf{Puzzle Generation Pipeline for GAP-3 and GAP-5 Datasets.} Both datasets follow the same four-step generation process: (a) Source images from The Metropolitan Museum of Art Open Access collection (CC0 1.0 Universal Public Domain Dedication); (b) Grid overlay defining puzzle piece boundaries; (c) VAE-based fragment generation creating irregular, archaeologically-realistic piece shapes; (d) Random shuffling producing the final puzzle configuration. The GAP-3 example uses \textit{Woman wearing floral hat, from the Novelties series (N228, Type 2) issued by Kinney Bros.} by Kinney Brothers Tobacco Company (1889), while the GAP-5 example uses \textit{Seated Figure} (700–1600 CE). Both artworks are from the Metropolitan Museum of Art's public domain collection, ensuring unrestricted use for research and publication.}
    \label{fig:puzzle_pipeline}
\end{figure*}

% \paragraph{\textit{Early days} of puzzle solving:}
% Initial computational approaches to puzzle assembly focused on developing compatibility metrics and greedy construction strategies. Classical methods evaluated piece relationships through hand-engineered features, measuring boundary dissimilarity or leveraging natural image statistics to determine adjacency likelihood. These techniques typically operated on small-scale datasets containing dozens of natural images, with pioneering datasets introduced by Cho \textit{et al.}~\cite{cho2010probabilistic} (20 puzzles), Pomeranz~\textit{et al.}~\cite{pomeranz2011fully} (26 puzzles), and Sholomon \textit{et al.}~\cite{sholomon2013genetic} (20 puzzles). 
% Assembly strategies generally proceeded in two phases: first constructing the puzzle border, then greedily filling interior positions using pairwise compatibility scores. More recently, optimization-based approaches have framed puzzle solving as a consistent labeling problem~\cite{khoroshiltseva2021jigsaw}, with subsequent work introducing multi-phase relaxation labeling techniques~\cite{vardi2023multi} to improve reconstruction quality on square puzzles.

\noindent\textbf{Classic optimization-based solvers:}
Traditional approaches to puzzle assembly primarily centered on integrating piece-compatibility metrics into various optimization schemes such as linear programming~\cite{yu2015solving}, greedy algorithms~\cite{pomeranz2011fully}, genetic optimization~\cite{sholomon2013genetic} and relaxation labeling~\cite{khoroshiltseva2021jigsaw,vardi2023multi}. Approximating adjacency between pieces was primarily determined by evaluating boundary similarity, image statistics or other hand-crafted features. Although these methods can effectively handle very large puzzle sizes, they are often evaluated using small datasets, containing a few dozen images.

\noindent\textbf{Early learning-based approaches.}
The advent of convolutional neural networks (CNNs) transformed puzzle solving by replacing manually designed features with trainable compatibility networks. Sholomon \etal pioneered this with DNN-Buddies~\cite{sholomon2016dnn}, using a Siamese network to learn edge compatibility and integrating its predictions into a classical greedy solver. Following with Deepzzle~\cite{paumard2020deepzzle}, Paumard \etal combined the predictions of a neighbor-detecting network with shortest-path optimization. They also introduced a dataset of 12,000 puzzles created from the Metropolitan Museum of Art (MET) images, featuring random linear gaps. Li \etal further expanded the learning paradigm with JigsawGAN~\cite{li2021jigsawgan}, merging permutation classification and adversarial generation to capture both edge cues and semantic context. To further handle piece erosion, \citet{bridger2020solving} employed adversarial discriminators to quantify the plausibility of inpainted regions between fragments. The TEN framework~\cite{rika2022ten} embedded entire fragments into a shared latent space to enhance piece adjacency prediction in the face of erosion. In GANzzle~\cite{talon2022ganzzle}, a complete mental image guides reconstruction via differentiable matching. This was extended in Ganzzle++~\cite{talon2025ganzzle++} by introducing global layout constraints with hierarchical assignment in a learned spatial-latent space.

\noindent\textbf{Recent learning-based architectures}
Recent advances in puzzle reassembly have been driven by the rise of transformer-based and generative architectures. \citet{chen2023jigsaw} utilized jigsaw puzzle solving as a pretext task for image classification in Vision Transformers (ViTs), while \citet{ren2023masked} enhanced this idea through masked-jigsaw positional embeddings. Later, \citet{heck2025solving} unified ViT encoders with permutation prediction heads to directly infer piece positions, followed by FCViT~\cite{kim2025solving}, which regresses over fragment coordinates rather than predicting a discrete permutation. Liu \etal introduced JPDVT~\cite{liu2024solving}, which employs a diffusion process to jointly place existing pieces while generating missing ones. Relatedly, Positional Diffusion~\cite{giuliari2024positional} formulates set ordering as a graph-based denoising task, while DiffAssemble~\cite{scarpellini2024diffassemble} provides a unified graph-diffusion framework for reassembly, while also supporting 3D. 
Several works tackled puzzle reconstruction through various reinforcement learning (RL) methods. SD$^2$RL~\cite{song2023siamese} applied deep Q-learning to optimize fragment swaps while introducing the popular JPwLEG dataset, containing 12,000 puzzles with fixed 44px and 12px gaps from MET Museum images. PDN-GA~\cite{song2023solving} integrated a genetic algorithm with a fragment cluster discriminant network. Later, ERL-MPP~\cite{song2025erlmpp} combined actor–critic reinforcement learning with evolutionary search and multi-head perception, while CEARI~\cite{song2025ceari} employs co-evolutionary agents to simultaneously reassemble and inpaint puzzles with gaps and missing pieces.
Most recently, multimodal solvers have emerged too. In particular, Xu and Liu introduced VLHSA~\cite{xu2025vlhsa}, leveraging vision-language hierarchical semantic alignment to enhance assembly performance on eroded puzzles, while \citet{elkin2025seq} demonstrated that language models can solve visual puzzles without utilizing the visual input except for tokenizing the pieces as discrete sequences.

Table~\ref{tab:2d_datasets} demonstrates the scale and erosion properties of prominent square jigsaw puzzles datasets, while Fig~\ref{fig:dataset_comparison} shows visual examples. Although other datasets of puzzles with non-square shapes exist, such as the RePAIR dataset~\cite{tsesmelis2024re} containing scanned archaeological fragments, the LSU puzzles repository~\cite{zhang14GMOD} containing both synthetic and scanned commercial/hand torn puzzles, and the GVC puzzles dataset~\cite{li2019hierarchical,le2019jigsawnet} containing synthetic puzzles made via random slicing curves, they are mostly limited in scope.

\begin{table*}[t]
\centering
\caption{Prominent square 2D puzzle solving datasets}
\label{tab:2d_datasets}
\begin{tabular}{ccccc}
\toprule
\textbf{Dataset} & \textbf{No. Puzzles} & \textbf{Erosion / Gap} & \textbf{Image Origin} & \textbf{Publicly Available?} \\
\midrule
Sholomon \textit{et al.} \cite{sholomon2013genetic} & 20 & X & Natural Images & Yes \\
Cho \textit{et al.} \cite{cho2010probabilistic} & 20 & X & Natural Images & Yes \\
Pomeranz~\textit{et al.}~\cite{pomeranz2011fully} & 26 & X & Natural Images & Yes \\
\midrule
Deepzzle~\cite{paumard2020deepzzle} & 12,000 & Random linear gap & MET Museum & No \\
JPwLEG-3~\cite{song2023siamese} & 12,000 & 44px gap & MET Museum & Yes \\
JPwLEG-5~\cite{song2023siamese} & 12,000 & 12px gap & MET Museum & Yes \\
% JPDVT~\cite{} & N/A* & Both with and without linear gap & ImageNet & No \\
\bottomrule
GAP-3 & 20,000 & Irregular erosion \& fragment shape & MET Museum & Yes \\
GAP-5 & 20,000 & Irregular erosion \& fragment shape & MET Museum & Yes \\
\end{tabular}
\end{table*}

\section{The GAP Datasets}
\label{sec:gap_datasets}

To address the dataset limitations identified in Section~\ref{sec:related_work}, we introduce GAP (Generated Archaeological Puzzles): two large-scale benchmark datasets (GAP-3 and GAP-5) featuring jigsaw puzzles with irregular fragment shapes, learned from archaeological data. Unlike existing benchmarks that maintain square piece geometries with fixed or linear gaps~\cite{paumard2020deepzzle, song2023siamese}, GAP employs fragment shapes generated from real archaeological distributions, creating variable, non-linear spacing that mirrors authentic erosion patterns. Each dataset contains 20,000 puzzles applied to diverse artwork images from the Metropolitan Museum of Art's Open Access collection~\cite{met_open_access}, providing visual diversity across cultures, time periods, and media types. By maintaining grid-based topology while introducing realistic geometric complexity, GAP bridges a gap between synthetic benchmarks and real-world archaeological applications while preserving compatibility with existing puzzle-solving methods.

\subsection{Fragment Shape Generator}
\label{sec:fragment_generator}

To generate realistic irregular fragments, we employ a Variational Autoencoder (VAE)~\cite{kingma2013auto} trained on 958 binary masks from the RePAIR dataset~\cite{tsesmelis2024re}, comprising real scanned archaeological fragments from the UNESCO World Heritage site of Pompeii.

\noindent\textbf{Architecture.} The VAE consists of: (1) an encoder with four convolutional layers (channels: 32, 64, 128, 256) reducing 128$\times$128 inputs to 256$\times$8$\times$8 features, (2) a 64-dimensional latent space with reparameterization via mean and log-variance projections, and (3) a symmetric decoder with four transposed convolutional layers upsampling to 128$\times$128 binary masks. We train for 44 epochs using Adam optimizer (lr=$10^{-4}$) with standard VAE loss balancing reconstruction (binary cross-entropy) and KL divergence regularization.

\noindent\textbf{Post-processing.} Generated masks are further processed in several elementary steps: (1) binarization at threshold 0.5 since the VAE outputs are continuous; (2) binary hole filling to eliminate interior voids, implemented by inverting the mask and propagating background pixels inward from the image boundary via iterative dilation, with holes identified as foreground regions unreachable from the border; (3) largest connected component selection; and (4) morphological closing using a disk-shaped structuring element (radius 2 pixels) to smooth the external boundary. These operations are made to ensure single, continuous fragments suitable for puzzle assembly.

\subsection{Dataset Construction}
\label{sec:dataset_construction}

\noindent\textbf{Image Source.} 
We curate 40,000 diverse images from The Metropolitan Museum of Art's Open Access collection~\cite{met_open_access}, spanning Asian Art, European Sculpture, Islamic Art, Photography, and other departments. Images represent diverse media types (paintings, ceramics, textiles, photographs), cultures, and temporal periods (2nd century BCE to 21st century CE), providing rich visual content and texture diversity. All images are licensed under CC0 1.0 Universal Public Domain Dedication, ensuring unrestricted use for research, distribution, and publication.

\noindent\textbf{Grid-Based Puzzle Generation.} 
To ensure compatibility with existing puzzle-solving methods and follow the most common benchmark dataset format~\cite{paumard2020deepzzle, song2023siamese}, we adopt a grid-based layout. For each puzzle: (1) randomly select a MET image and resize to canvas size (384$\times$384 for GAP-3, 640$\times$640 for GAP-5), (2) overlay a regular $n \times n$ grid ($3\times3$ or $5\times5$), (3) generate VAE fragment masks positioned at grid cell centers, (4) extract textured fragments by applying masks to the image, and (5) record ground truth (grid positions, piece IDs, complete reference image). This yields 9-piece (GAP-3) and 25-piece (GAP-5) puzzles with irregular, archaeologically-inspired shapes while maintaining the structured topology that most algorithms expect. Train/validation/test splits (70/15/15) ensure no image overlap across splits, with GAP-3 and GAP-5 using entirely separate image sets to allow both independent and combined multi-scale utilization/evaluation.

\subsection{Geometric Validation of Generated Fragments}
\label{sec:fragment_analysis}

To validate geometric fidelity, we compare 958 VAE-generated fragments against 958 real archaeological fragments from RePAIR~\cite{tsesmelis2024re} across eight geometric features: area, perimeter, aspect ratio, solidity, circularity, compactness, vertices, and concavities. See full description of these features in Supp.~\ref{sec:supp_validation}

\noindent\textbf{Core Shape Fidelity Features:} 
Generated fragments preserve fundamental geometric properties with high accuracy: mean area differs by $<1\%$ (10,617 vs. 10,716 px$^2$), aspect ratio by 3\%, and solidity by 2\%. These small differences confirm that GAP fragments maintain the size distribution and shape proportions of real archaeological fragments.

\noindent\textbf{Edge Complexity Features:} 
As expected from VAE reconstruction and post-processing, edge features show moderate differences: perimeter (12\% difference), circularity (18\%), vertices (22\%), and concavities (19\%). These differences reflect the VAE's learned smoothing characteristics and morphological post-processing operations, which produce cleaner boundaries than natural fracture processes while maintaining realistic edge irregularity.

\begin{figure}[t]
    \centering
    \includegraphics[width=\linewidth]{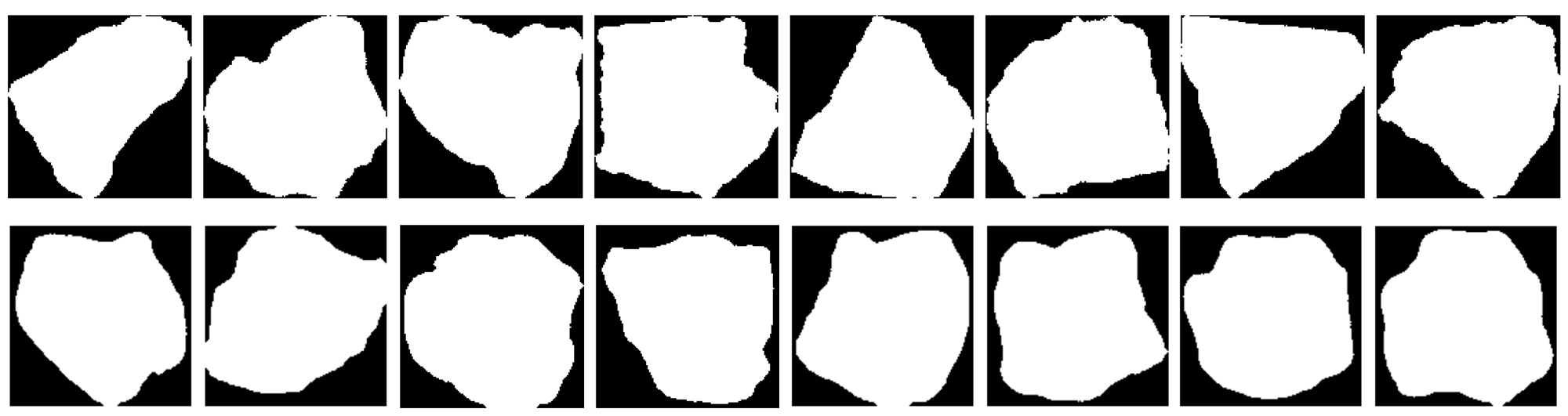}
    \caption{Qualitative comparison: real archaeological fragments from RePAIR (top) vs. Generated fragments (bottom). Synthetic fragments exhibit similar irregular shapes and edge complexity.}
    \label{fig:vae_samples}
\end{figure}

\begin{figure}[t]
    \centering
    \includegraphics[width=0.95\linewidth]{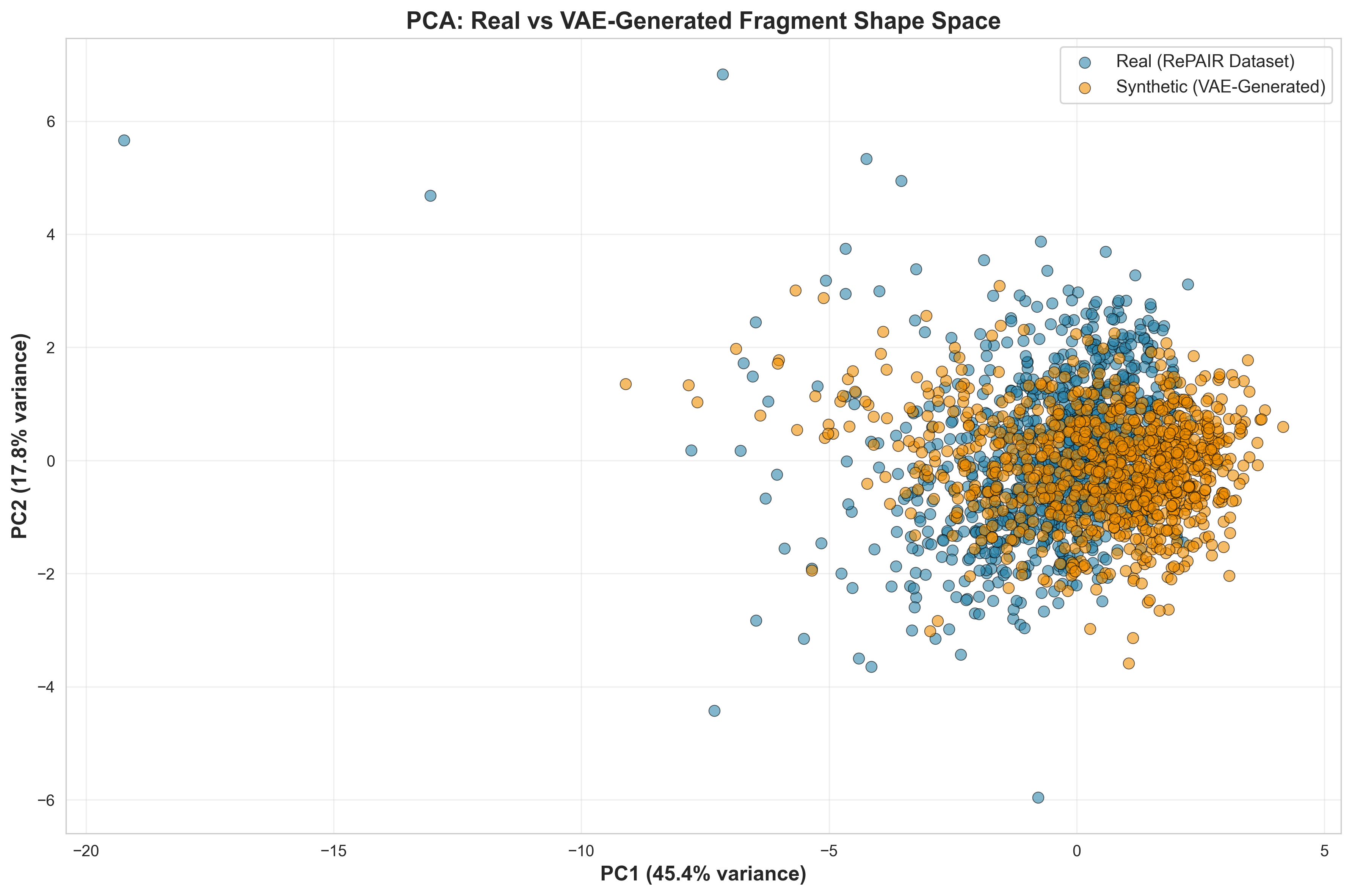}
    \caption{PCA embedding of geometric features (63.2\% variance explained). Real archaeological fragments (blue) and VAE-generated fragments (orange) show substantial distributional overlap, confirming the VAE captures diverse fragment morphologies without mode collapse.}
    \label{fig:vae_embedding}
\end{figure}

\noindent\textbf{Distribution Coverage:} Figure~\ref{fig:vae_samples} shows qualitative similarity between real and synthetic fragments. Figure~\ref{fig:vae_embedding} presents PCA projections (63.2\% variance), revealing substantial distributional overlap with no mode collapse. See complete statistical analysis in Supp.~\ref{sec:supp_validation}.

While authentic archaeological materials remain the gold standard for final validation, the GAP datasets provide a crucial, controlled, and yet highly realistic testbed that enables systematic algorithm development and comparative evaluation at scale. The complete datasets (40,000 puzzles total), trained VAE model, evaluation scripts, and baseline implementations will be made publicly available upon acceptance to facilitate reproducible research and standardized algorithm comparison.

\section{PuzzleFlow - Solving puzzles with Flow Matching}
\label{sec:method}

We formulate jigsaw puzzle reassembly as a \emph{permutation learning} problem using discrete flow matching. Our approach leverages pretrained ViT to model piece relationships, enabling end-to-end differentiable learning of valid permutations.

\subsection{Problem Formulation}

Given $N$ shuffled puzzle pieces $\mathcal{X} = \{x_1, \ldots, x_N\}$ from a $k \times k$ grid (where $N = k^2$), we seek the permutation $\pi^* \in \mathcal{S}_N$ (the symmetric group of all permutations of $N$ elements)  that maps each piece to its ground truth position. 
We thus pursue a model $f_{\theta}: \mathcal{X} \rightarrow \mathcal{S}_N$ that maximizes
\begin{equation}
\pi^* = \arg\max{\pi \in \mathcal{S}_N} , p{\theta}(\pi \mid \mathcal{X}) ,
\label{eq:objective}
\end{equation}
where $\theta$ denotes the learnable model parameters. (Later, $t$ denotes flow time; we use distinct notation to avoid confusion.)
Clearly, the combinatorial search space (of $N!$ possible configurations) and the discrete output structure pose significant optimization challenges.

\subsection{Discrete Flow Matching}

To pursue a learned model, we adopt flow matching~\cite{lipman2022flow} extended to discrete permutations~\cite{alfonso2023generative}. Instead of directly predicting $\pi^*$, we model a time-dependent distribution $p_t(\pi_t \mid \mathcal{X})$ where $t \in [0, 1]$:
\begin{itemize}
    \item At $t=0$: $\pi_0 \sim \text{Uniform}(\mathcal{S}_N)$ (random permutation)
    \item At $t=1$: $\pi_1 = \pi^*$ (ground truth)
    \item For $t \in (0, 1)$: $\pi_t$ represents interpolated state
\end{itemize}

\noindent\textbf{Stochastic Interpolation.} At time $t$, each piece $i$ is assigned to target position $\pi_1^{(i)}$ with probability $\alpha(t) = t$ (linear schedule):
\begin{equation}
\pi_t^{(i)} = \begin{cases} 
\pi_1^{(i)} & \text{with probability } t \\
\pi_0^{(i)} & \text{with probability } 1 - t
\end{cases}
\end{equation}

\noindent\textbf{Training Objective.} The model predicts target positions conditioned on current state $\pi_t$ and time $t$:
\begin{equation}
\mathcal{L}_{\text{CFM}} = \mathbb{E}_{t, \pi_0, \pi_t} \left[ -\sum_{i=1}^N \log p_\theta(\pi_1^{(i)} \mid x_i, \pi_t, t) \right]
\end{equation}
This enables learning incremental refinements rather than single-shot predictions.

\subsection{Architecture}

\noindent\textbf{Visual Encoding.} We employ ViT-Base~\cite{dosovitskiy2020image} pretrained on ImageNet-21K as our feature extractor. Puzzle pieces are stored as 128×128px RGBA images where the alpha channel encodes the mask of the irregular fragment shape.

\noindent\textbf{RGBA Adaptation.} To handle irregular fragments in GAP, we employ a learned 1×1 convolutional layer that projects RGBA $\to$ RGB before ViT encoding. Unlike square-piece methods that can simply discard the alpha channel, this learned projection adaptively combines all four channels to preserve fragment shape information encoded in the alpha mask. This design choice is critical for irregular puzzles where boundary geometry provides essential spatial cues. Indeed, ablation studies (§\ref{subsec:ablations}) demonstrate that naive alpha-dropping causes severe performance degradation.

Following channel adaptation, pieces are resized to $224 \times 224$px via bilinear interpolation and processed through the pretrained encoder. This encoder (ViT-Base) tokenizes each image into $14 \times 14 = 196$ patches, plus one \texttt{[CLS]} token. For each piece, we use the output vector of the \texttt{[CLS]} token as its feature representation: $h_i \in \mathbb{R}^{768}$. The 768-dimensionality corresponds to the hidden size of the ViT-Base model used, and represents a global summary embedding for the input piece.

\noindent\textbf{Conditioning.} We augment visual features with learned embeddings that encode the flow state. All embeddings share the same 768-dimensional space and are combined via residual addition:
\begin{itemize}[leftmargin=*, noitemsep, topsep=0pt]
    \item \textbf{Position:} Current position index $p_i \in \{0, \ldots, N-1\}$ is encoded via learned lookup table $\mathbf{E}_{\text{pos}} \in \mathbb{R}^{N \times 768}$, yielding position embedding $\mathbf{e}_{\text{pos}}(p_i) \in \mathbb{R}^{768}$.
    \item \textbf{Time:} Flow time $t \in [0,1]$ is encoded via 192-dimensional sinusoidal embedding~\cite{vaswani2017attention} followed by a two-layer MultiLayer Perceptron (MLP) with SiLU activation, producing time embedding $\mathbf{e}_{\text{time}}(t) \in \mathbb{R}^{768}$.
\end{itemize}

The conditioned representation for piece $i$ is computed as:
\begin{equation}
z_i = h_i + \mathbf{e}_{\text{pos}}(p_i) + \mathbf{e}_{\text{time}}(t) \in \mathbb{R}^{768}
\end{equation}
jointly encoding 
\emph{what} each piece looks like, 
\emph{where} it currently is, and 
\emph{when} are we in the flow process.

\noindent\textbf{Relational Reasoning.} We apply $L=4$ transformer encoder layers with pre-normalization (12 attention heads, hidden dimension 768, feedforward dimension 3072) to the sequence $\{z_1, \ldots, z_N\}$. These layers enable each piece representation to attend to all other pieces, learning holistic visual relationships across the entire puzzle configuration.

\noindent\textbf{Output Prediction.} An MLP head (768 $\to$ 3072 $\to$ $N$) outputs position logits $\ell_i \in \mathbb{R}^N$ for each piece. Softmax yields probabilities:
\begin{equation}
p_\theta(\pi_1^{(i)} = j \mid x_i, \pi_t, t) = \frac{\exp(\ell_i[j])}{\sum_{j'} \exp(\ell_i[j'])}, j'=1..N
\end{equation}

\subsection{Training Details}
Training employed AdamW optimizer~\cite{loshchilov2017decoupled} with learning rate $10^{-5}$, weight decay $0.01$, OneCycleLR schedule with 10\% warmup, dropout $p=0.1$, and automatic mixed precision (FP16). Training requires 30 epochs with batch size 8 on RTX4090 GPUs. See full hyperparameters in Supp.~\ref{sec:implementation_supplementary}.

\subsection{Inference}
At test time, we perform iterative refinement starting from random permutation $\pi_0 \sim \text{Uniform}(\mathcal{S}_N)$. For $S=20$ steps, at each timestep $t = s/S$ we:
\begin{enumerate}[leftmargin=*, noitemsep, topsep=0pt]
    \item Compute logits $\ell \leftarrow f_\theta(\mathcal{X}, \pi_{s-1}, t)$
    \item Update via greedy assignment: $\pi_s^{(i)} = \argmax_{j \in \mathcal{P}_{\text{avail}}} \ell_i[j]$
\end{enumerate}
where $\mathcal{P}_{\text{avail}}$ is the set of unassigned positions. This runs in $O(N^2)$ time, far more efficient than exhaustive search ($O(N!)$) or Hungarian matching ($O(N^3)$), commonly used in recent puzzle-solving frameworks.

\section{Experiments}
\label{sec:experiments}

We present comprehensive experiments with two objectives: (1) \textbf{benchmarking GAP as a challenging testbed} featuring irregular, eroded fragments that better reflect archaeological scenarios than existing square-piece datasets, and (2) \textbf{validating PuzzleFlow's design} through systematic evaluation showing substantial improvements over state-of-the-art (SOTA) methods and rigorous ablation studies.

\subsection{Experimental Setup}

\noindent\textbf{Datasets.} We evaluate on GAP-3 and GAP-5 (20,000 puzzles each, Section~\ref{sec:gap_datasets}) using standard 70/15/15 train/validation/test splits with 3,000 test puzzles per dataset. \\
\noindent\textbf{Evaluation Metrics.}
We adopt \textbf{Perfect Accuracy (PA)}, percentage of completely solved puzzles, and \textbf{Absolute Accuracy (AA)}, percentage of correctly placed pieces, as established in prior work~\cite{paumard2020deepzzle, song2023siamese, liu2024solving}. However, these metrics measure only absolute positional correctness and cannot distinguish between predictions that preserve local spatial structure versus random permutations.

\noindent In addition, we employ \textbf{Spatial Relationship Accuracy (SRA)} to capture whether models learn coherent spatial relationships. Extending Song~\textit{et al.}'s~\cite{song2023siamese} directional metrics and earlier neighborhood metrics~\cite{cho2010probabilistic,gur2017square}, SRA measures the fraction of ground-truth neighbor pairs that remain neighbors \emph{in the same relative spatial configuration} in the prediction. For example, if pieces A and B are horizontal neighbors in the ground truth (A left of B), SRA counts this as preserved only if they remain horizontal neighbors in the prediction (A still left of B), not if they become vertical neighbors or are separated. Formally:
\begin{equation}
    \text{SRA} = \frac{1}{M} \sum_{i=1}^{M} \frac{|\{(u,v,d) \in \mathcal{N} : \text{rel}(\mathbf{p}_i, u, v) = d\}|}{|\mathcal{N}|}
\end{equation}
where $\mathcal{N}$ contains all neighbor pairs in the $g \times g$ grid with their directional relationships $d \in \{\text{left, right, up, down}\}$, and $\text{rel}(\mathbf{p}, u, v)$ checks if pieces at ground-truth positions $u$ and $v$ maintain the same directional relationship $d$ under predicted permutation $\mathbf{p}$. High SRA with moderate AA indicates learned local structure despite global placement errors, while low SRA suggests random-like predictions.

% \begin{table}[t]
\begin{table*}[t]
\centering
\caption{\textbf{Main results on GAP datasets.} Perfect Accuracy (PA), Absolute Accuracy (AA), and Spatial Relationship Accuracy (SRA) on test sets. Best in \textbf{bold}, second-best \underline{underlined}.}
\label{tab:main_results}
% \resizebox{\columnwidth}{!}{%
\begin{tabular}{l|ccc|ccc}
\toprule
& \multicolumn{3}{c|}{\textbf{GAP-3 (3$\times$3)}} & \multicolumn{3}{c}{\textbf{GAP-5 (5$\times$5)}} \\
\textbf{Method} & PA (\%) $\uparrow$ & AA (\%) $\uparrow$ & SRA (\%) $\uparrow$ & PA (\%) $\uparrow$ & AA (\%) $\uparrow$ & SRA (\%) $\uparrow$ \\
\midrule
\textit{Classical} \\
Greedy~\cite{pomeranz2011fully} & 0.0 & 11.6 & 8.6 & 0.0 & 4.1 & 3.7 \\
GA~\cite{sholomon2013genetic}& 0.0 & 11.1 & 8.5 & 0.0 & 11.1 & 8.5 \\
\midrule
\textit{Deep Learning} \\
JigsawGAN~\cite{li2021jigsawgan} & 4.6 & 45.3 & 35.9 & 0.0 & 18.0 & 12.0 \\
DiffAssemble~\cite{scarpellini2024diffassemble} & 16.4 & 50.5 & 43.4 & 0.0 & \underline{21.9} & \underline{14.7} \\
JPDVT~\cite{liu2024solving} & 0.0 & 11.2 & 8.4 & 0.0 & 3.9 & 3.2 \\
\citet{elkin2025seq} & 0.0 & 14.8 & 9.9 & 0.0 & 7.8 & 4.5 \\
FCViT~\cite{kim2025solving} & \underline{25.2} & \underline{60.7} & \underline{47.6} & 0.0 & 20.4 & 13.8 \\
\midrule
\textbf{PuzzleFlow (ours)} & \textbf{28.5} & \textbf{62.9} & \textbf{55.7} & \textbf{0.3} & \textbf{29.1} & \textbf{19.8} \\
\bottomrule
\end{tabular}
% }
\end{table*}
% \end{table}

\noindent\textbf{Baseline Methods.}
Since several recent methods did not release their implementations publicly,  including them in a comparison was infeasible. We used the strongest available baselines including top-performing published results on JPwLEG, as well as implementing baselines based on classic non-learning based approaches. In general, we compare against seven methods: \textbf{Greedy Solver} based on Pomeranz~\textit{et al.}~\cite{pomeranz2011fully} using edge compatibility, \textbf{Genetic Algorithm} inspired by Sholomon~\textit{et al.}~\cite{sholomon2013genetic} with priority-based encoding, and five prominent deep learning methods that achieved top performance on JPwLEG benchmarks: \textbf{FCViT}~\cite{kim2025solving} performing continuous coordinate regression, the diffusion based \textbf{JPDVT}~\cite{liu2024solving} and \textbf{DiffAssemble}~\cite{scarpellini2024diffassemble}, GAN based \textbf{JigsawGAN}~\cite{li2021jigsawgan}, and the recent \textbf{PuzLM}~\cite{elkin2025seq} processing visual tokens as sequences. All deep learning methods use comparable capacity when possible (ViT-Base, approximately 85-124M parameters) and are retrained on GAP with similar budgets (30 epochs).

In terms of the comparison, it is worth noting most deep-learning approaches were originally designed for square RGB pieces on datasets like JPwLEG~\cite{song2023siamese} where alpha channels are unnecessary. When applied to GAP's RGBA fragments, these methods naturally process only RGB channels. In order to run them on GAP, we adapted the RGBA images to RGB as a pre-processing step with normalization to imagenet values. 
% Ablation studies (§\ref{subsec:ablations}) suggest this reflects a fundamental difference in problem settings rather than architectural limitations.

% \paragraph{Training Configuration.}
% PuzzleFlow uses ViT-B/16 pretrained on ImageNet-21k with $L=4$ additional transformer layers (12 heads, 1024 hidden dim). We train with AdamW ($\text{lr}=10^{-5}$, weight decay $0.01$), OneCycleLR schedule, batch size 8 for 30 epochs (GAP-3) or 50 epochs (GAP-5), requiring $\sim$18 hours on RTX4090. Inference uses $S=20$ iterative refinement steps. Full details in supplementary materials.

\subsection{Main Results}
\label{subsec:main_results}

Table~\ref{tab:main_results} presents results on GAP-3 and GAP-5. PuzzleFlow substantially outperforms all baselines, validating that our architectural design, combining flow matching with fine-tuned visual features, relational reasoning and iterative refinement, enables effective reconstruction on irregular, eroded fragments. Importantly, it also establishes the imperative of GAP as a benchmark dataset, as it clearly challenges existing and future methods much more significantly.

% \begin{table*}[t]
% \centering
% \caption{\textbf{Main results on GAP datasets.} Perfect Accuracy (PA), Absolute Accuracy (AA), and Spatial Relationship Accuracy (SRA) on test sets. Best in \textbf{bold}, second-best \underline{underlined}.}
% \label{tab:main_results}
% \begin{tabular}{l|ccc|ccc}
% \toprule
% & \multicolumn{3}{c|}{\textbf{GAP-3 (3$\times$3)}} & \multicolumn{3}{c}{\textbf{GAP-5 (5$\times$5)}} \\
% \textbf{Method} & PA (\%) $\uparrow$ & AA (\%) $\uparrow$ & SRA (\%) $\uparrow$ & PA (\%) $\uparrow$ & AA (\%) $\uparrow$ & SRA (\%) $\uparrow$ \\
% \midrule
% \textit{Classical} \\
% Greedy~\cite{pomeranz2011fully} & 0.0 & 11.4 & 8.6 & 0.0 & 4.0 & \underline{3.7} \\
% GA~\cite{sholomon2013genetic}& 0.0 & 11.0 & 8.4 & 0.0 & 3.9 & 3.4 \\
% \midrule
% \textit{Deep Learning} \\
% FCViT~\cite{kim2025solving} & 0.0 & \underline{14.0} & \underline{11.0} & 0.0 & 4.0 & 3.0 \\
% JPDVT~\cite{liu2024solving} & 0.0 & 11.2 & 8.5 & 0.0 & 4.1 & 3.4 \\
% PuzLM~\cite{elkin2025seq} & 0.0 & 13.8 & 9.1 & 0.0 & \underline{7.6} & \underline{4.4} \\
% \midrule
% \textbf{PuzzleFlow (ours)} & \textbf{26.2} & \textbf{60.8} & \textbf{53.5} & \textbf{0.2} & \textbf{28.8} & \textbf{19.8} \\
% \bottomrule
% \end{tabular}
% \end{table*}

\noindent\textbf{GAP-3:}
Classical approaches (Greedy, GA) and several deep learning methods (JPDVT, PuzLM) achieve 0\% PA and near-random AA (11-15\%), confirming that GAP's irregular geometries and edge erosion break assumptions underlying boundary-matching and local-feature methods. In contrast, some methods demonstrate meaningful performance: JigsawGAN (4.6\% PA, 45.3\% AA), DiffAssemble (16.4\% PA, 50.5\% AA), FCViT (25.2\% PA, 60.7\% AA). PuzzleFlow achieves the highest results with 28.5\% PA and 62.9\% AA. More notably, PuzzleFlow's SRA of 55.7\% substantially exceeds the second-best DiffAssemble's 43.4\% (+12.3 points) and FCViT's 47.6\% (+8.1 points), indicating that our architecture captures significantly better spatial coherence. These results establishes GAP as a timely, challenging yet tractable benchmark; difficult enough for existing methods, but solvable with appropriate architectural design. The remaining $\sim$71\% unsolved puzzles provide substantial headroom for future work.

% All baseline methods achieve 0\% perfect accuracy, failing to solve even a single puzzle among 3,000 test cases. This validates GAP's challenge: irregular geometries and edge erosion break assumptions underlying existing methods. Classical approaches (Greedy, GA) achieve only 11.0-11.4\% AA, barely above random (11.1\%), as boundary-matching fails when edges are degraded. Deep learning baselines fare marginally better (11.2-14.0\% AA).
% % , but FCViT's coordinate regression cannot handle irregular shapes, while JPDVT's local features lack global context. 
% % Notably, FCViT reports XX\% PA on square-piece benchmarks~\cite{kim2025solving} but 0\% on GAP-3 despite identical puzzle size (9 pieces), quantifying GAP's substantially greater difficulty.

% Unlike the prior art, PuzzleFlow achieves 26.2\% PA and 60.8\% AA, representing absolute gains of +26.2 and +46.8 points over baselines. The 53.5\% SRA, substantially exceeding baseline SRA (8-11\%), demonstrates learned spatial structure rather than random predictions. This establishes GAP as a timely, challenging yet tractable benchmark; difficult enough for existing methods, but solvable with appropriate architectural design. The remaining 74\% unsolved (or partially) puzzles provide substantial headroom for future algorithmic development.

\noindent\textbf{GAP-5:}
With 25 pieces, the combinatorial complexity increases dramatically ($25! \approx 1.55 \times 10^{25}$ vs. $9! \approx 3.6 \times 10^5$). While several baselines degrade to near-random levels, three methods maintain meaningful performance in this more challenging setting: JigsawGAN (18.0\% AA), FCViT (20.4\% AA), and DiffAssemble (21.9\% AA, 14.7\% SRA), demonstrating that diverse learning-based approaches can partially generalize to larger configurations.
PuzzleFlow achieves the best results across all metrics: 0.3\% PA, 29.1\% AA, and 19.8\% SRA, outperforming the second-best DiffAssemble by +7.2 AA and +5.1 SRA points. The performance gap between PuzzleFlow and the strongest baselines widens from GAP-3 to GAP-5. This validates that our architecture enables holistic visual reasoning that partially survives the transition to larger configurations, whereas methods relying on local boundary features degrade to random performance.

% With 25 pieces, the combinatorial complexity increases dramatically ($25! \approx 1.55 \times 10^{25}$ vs. $9! \approx 3.6 \times 10^5$). Baseline and prior art performance degrades to near-random levels: 3.9-4.1\% AA approximates expected random accuracy (4\%), and SRA drops to 3.0-3.7\%. Even learned representations provide no advantage over hand-crafted features when both fail catastrophically.

% PuzzleFlow maintains meaningful performance: 0.2\% PA (i.e., despite the complexity, several puzzles were still solved perfectly), 28.8\% AA, 19.8\% SRA. While perfect reconstruction becomes rare, the model correctly places nearly one-third of all pieces, 7x better than baselines, and preserves 5x more spatial relationships (19.8\% vs. 3.0-3.7\%). This validates that our architecture enables holistic visual reasoning that partially survives the transition to larger configurations, whereas methods relying on local boundary features degrade to random performance.

\subsection{Ablation Studies}
\label{subsec:ablations}

We conduct systematic ablations on GAP-3 to validate design choices. All variants use identical training protocols to isolate architectural effects. Table~\ref{tab:ablations} summarizes results.

\begin{table}[t]
\centering
\caption{\textbf{Ablation studies on GAP-3.} $\Delta$ shows difference from full model.}
\label{tab:ablations}
\resizebox{\columnwidth}{!}{%
\begin{tabular}{l|ccc|ccc}
\toprule
\textbf{Variant} & \textbf{PA} & \textbf{AA} & \textbf{SRA} & \textbf{$\Delta$PA} & \textbf{$\Delta$AA} & \textbf{$\Delta$SRA} \\
\midrule
\textit{Core Framework} \\
Direct Prediction      & 22.6 & 57.9 & 50.0 & -5.9 & -5.0 & -5.7 \\
Frozen ViT             & 7.4 & 42.2 & 34.5 & -21.1 & -20.7 & -21.2 \\
\midrule
\textit{Architecture Depth} \\
0 Layers               & 10.1 & 45.1 & 35.3 & -18.4 & -17.8 & -20.4 \\
2 Layers               & 23.5 & 58.8 & 50.6 & -5.0 & -4.1 & -5.1 \\
6 Layers               & 24.7 & 59.5 & 52.2 & -3.8 & -3.4 & -3.5 \\
\midrule
\textit{RGBA Adaptation} \\
Fixed Slicing (RGB-only) & 9.2 & 44.4 & 34.6 & -19.3 & -18.5 & -21.1 \\
\midrule
\textbf{Full Model} & \textbf{28.5} & \textbf{62.9} & \textbf{55.7} & -- & -- & -- \\
\bottomrule
\end{tabular}%
}
\end{table}

% \begin{table}[t]
% \centering
% \caption{\textbf{Ablation studies on GAP-3.} $\Delta$ shows difference from full model.}
% \label{tab:ablations}
% \resizebox{\columnwidth}{!}{%
% \begin{tabular}{l|ccc|ccc}
% \toprule
% \textbf{Variant} & \textbf{PA} & \textbf{AA} & \textbf{SRA} & \textbf{$\Delta$PA} & \textbf{$\Delta$AA} & \textbf{$\Delta$SRA} \\
% \midrule
% \textit{Core Framework} \\
% Direct Prediction      & 24.7 & 58.7 & 51.6 & -1.5 & -2.1 & -1.9 \\
% Frozen ViT             & 5.1 & 39.6 & 30.9 & -21.1 & -21.2 & -22.6 \\
% \midrule
% \textit{Architecture Depth} \\
% 0 Layers               & 10.7 & 45.9 & 35.8 & -15.5 & -14.9 & -17.7 \\
% 2 Layers               & 24.8 & 60.2 & 52.6 & -1.4 & -0.6 & -0.9 \\
% 6 Layers               & 27.1 & 61.6 & 54.4 & +0.9 & +0.8 & +0.9 \\
% \midrule
% \textit{RGBA Adaptation} \\
% Fixed Slicing (RGB-only) & 2.3 & 31.8 & 22.0 & -23.9 & -29.0 & -31.5 \\
% \midrule
% \textbf{Full Model} & \textbf{26.2} & \textbf{60.8} & \textbf{53.5} & -- & -- & -- \\
% \bottomrule
% \end{tabular}%
% }
% \end{table}

\noindent\textbf{Flow Matching vs. Direct Prediction:}
Replacing iterative flow matching with single-shot cross-entropy prediction yields 22.6\% PA (-5.9 points), showing flow matching provides consistent improvements. While more sophisticated inference (e.g., ancestral sampling) could increase gains even further, consistent improvements across all 3 metrics validate that iterative refinement helps resolve ambiguities, particularly for pieces with weak visual anchors.

\noindent\textbf{ViT Fine-Tuning:}
Freezing the pretrained ViT leads to a major drop to 7.4\% PA (-21.1 points), the largest among all ablations. This establishes fine-tuning as critical for irregular fragments. ImageNet features require adaptation to learn cross-boundary continuity, erosion robustness, and global coherence patterns specific to archaeological puzzles.

\noindent\textbf{Architecture Depth:}
Varying task-specific layers ($L \in \{0, 2, 4, 6\}$) reveals clear trends: $L=0$ achieves only 10.1\% PA, confirming pretrained features alone are insufficient. Performance jumps to 23.5\% at $L=2$ (+13.4 points), then plateaus near $L=4$ (+5.0 points). $L=6$ do not show gains (24.7\% PA, -3.8 points), suggesting deeper architectures do not necessarily improve further. We choose $L=4$ for the best balance between accuracy and computational efficiency.

\noindent\textbf{RGBA Adaptation for Irregular Fragments:}
We assess the necessity of our learned RGBA to RGB projection for handling irregular fragments with explicit shape masks. To this end, we perform an ablation at inference by replacing our projection with direct RGB slicing, which discards the alpha channel post training and reflects standard practice in square-piece puzzle methods. As shown in Table~\ref{tab:ablations}, removing the alpha channel leads to a pronounced reduction in accuracy across all metrics, underscoring that shape information encoded by the alpha mask is vital for successful assembly of irregular archaeological fragments. 
Rather than granting an unfair advantage, our treatment represents an essential adaptation to a fundamentally different task. In other words, we conclude that methods designed for regular, square fragments do not need or rely on more explicit shape encoding, but irregular puzzles require it for precise reconstruction. \\
% This ablation highlights how modeling shape is critical in complex, real-world puzzle settings.
\noindent\textbf{Ablation Summary:}
Ablations reveal four key insights: (1) Fine-tuning dominates (+21.1 PA points), dwarfing other factors. Future work should thus prioritize transfer learning strategies; (2) RGBA adaptation is critical for irregular fragments (+19.3 PA points), but reflects problem-specific necessity rather than unfair advantage. It is not controversial to include readily available shape information in problems where shape plays an important role; (3) Flow matching provides meaningful improvement (+5.9 points). Consistent improvements indeed validate the approach, but better inference algorithms may unlock even larger gains; (4) Moderate depth suffices, where $L=4$ balances capacity and efficiency. The \textit{combination} of iterative flow matching, fine-tuning, shape representation, and appropriate architectural depth drives the obtained results, outperformaing the prior art significantly. At the same time, the remaining headroom (74\% PA gap of partially or fully unsolved puzzles) ensures that GAP serves as a valuable ongoing benchmark for the community. \\
\noindent To verify generalization, we further evaluate PuzzleFlow on standard square-piece benchmarks. While not specialized for square settings, PuzzleFlow achieves competitive results on JPwLEG-3 and JPwLEG-5 and remains consistent with its performance on GAP dataset. Detailed results and comparisons are provided in Supp.~\ref{sup:jpwleg_validation}.\\
Supp.~\ref{fig:qualitative_results} demonstrates representative examples of PuzzleFlow solving GAP puzzles. Note that in some cases generally low evaluation scores could be explained by relatively similar pieces, mistakenly assigned to the correspondent position.

\section{Conclusions}
This work proposes an important bridge to a longstanding disparity between academic jigsaw puzzle benchmarks and the complex realities of archaeological reconstruction. First, the GAP datasets introduce large-scale, systematically generated puzzles featuring authentic, irregular fragment shapes and realistic erosion patterns, learned from and closely matching the challenges faced in cultural heritage applications. Second, we present PuzzleFlow as a complimentary solver, leveraging ViTs and discrete flow matching, which learns holistic visual relationships across entire fragments rather than relying solely on edge continuity. Extensive experiments demonstrate that PuzzleFlow consistently outperforms adapted baselines on GAP, with ablation studies confirming the value of fine-tuned features, architectural depth, and explicit fragment shape representation. Despite these gains, we acknowledge that PuzzleFlow shares common limitations with most recent learning-based architectures regarding its scalability to very large fragment counts and its reliance on a structured grid topology.
Looking forward, extending PuzzleFlow to handle missing fragments, irregular spatial arrangements beyond grid topologies, and integration of physical constraints represents promising directions for practical archaeological applications. We hope that GAP datasets and our open-source implementation will catalyze further research bridging computer vision and digital heritage preservation. All code and datasets are publicly available.\\
\noindent\textbf{Ethical Statement.} Ethical considerations regarding the use of cultural heritage data and the generation of synthetic fragments are discussed in Supp.~\ref{sup:ethical_statement}.

\section*{Acknowledgments}
% \subsubsection{\ackname} 
This work has been funded in part by the European Union’s Horizon 2020 research and innovation programme under grant agreement No 964854 (the RePAIR project). The authors also acknowledge the use of generative AI tools for technical assistance in code implementation and linguistic refinement. 

% We also acknowledge the use of large language models, including Google Gemini, Claude (Anthropic), GitHub Copilot, and Perplexity, for assistance in code implementation and manuscript revision. 

% \section{Methods}

% \input{sec/0_abstract}    
% \input{sec/1_intro}
% \input{sec/2_formatting}
% \input{sec/3_finalcopy}
{
    \small
    \bibliographystyle{ieeenat_fullname}
    \bibliography{cleab_bib}
}

\clearpage
\setcounter{page}{1}
\maketitlesupplementary

%%%%%%%%%%%%%%%%%%%%%%%%%%%%%%%%%%%%%%%%%%%%%%%%%%%%%%%%%%%%%%%%%%%%%%%%%%%%%%%%
\section{Introduction}
\label{sec:supp_intro}
%%%%%%%%%%%%%%%%%%%%%%%%%%%%%%%%%%%%%%%%%%%%%%%%%%%%%%%%%%%%%%%%%%%%%%%%%%%%%%%%

This supplementary material provides comprehensive technical documentation for all components of our work. We organize the content into three main sections: (1) the GAP dataset generation pipeline and statistical validation (Section~\ref{sec:gap_supplementary}), (2) complete implementation details and training configurations for PuzzleFlow and all baseline methods (Section~\ref{sec:implementation_supplementary}), and (3) qualitative results showcasing puzzle reconstructions (Section~\ref{sec:qualitative_results}). Together, these sections enable full reproduction of our experiments and provide deeper insights into our methodological choices.

%%%%%%%%%%%%%%%%%%%%%%%%%%%%%%%%%%%%%%%%%%%%%%%%%%%%%%%%%%%%%%%%%%%%%%%%%%%%%%%%
\section{GAP Dataset: Generation and Validation}
\label{sec:gap_supplementary}
%%%%%%%%%%%%%%%%%%%%%%%%%%%%%%%%%%%%%%%%%%%%%%%%%%%%%%%%%%%%%%%%%%%%%%%%%%%%%%%%

This section details the complete pipeline for generating the GAP (Generated Archaeological-fragments Puzzles) datasets, including the fragment generator architecture, training procedure, source data collection, and comprehensive statistical validation against real archaeological fragments.

%%%%%%%%%%%%%%%%%%%%%%%%%%%%%%%%%%%%%%%%%%%%%%%%%%%%%%%%%%%%%%%%%%%%%%%%%%%%%%%%
\subsection{Fragment Generator Architecture}
\label{sec:supp_vae}
%%%%%%%%%%%%%%%%%%%%%%%%%%%%%%%%%%%%%%%%%%%%%%%%%%%%%%%%%%%%%%%%%%%%%%%%%%%%%%%%
\begin{figure*}[ht]
    \centering
    \includegraphics[width=\linewidth]{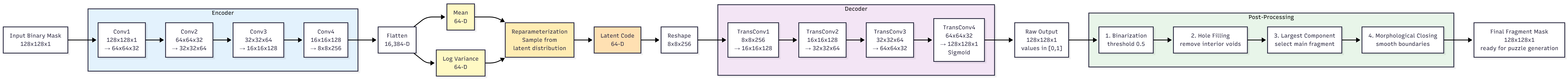}
    \caption{\textbf{Fragment Generator Architecture.} Our VAE encodes 128×128 binary fragment masks through four convolutional layers into a 64-dimensional latent space, then reconstructs synthetic fragments via transposed convolutions. The reparameterization trick enables sampling diverse fragments during training while maintaining archaeological realism.}
    \label{fig:vae_architecture}
\end{figure*}

Our fragment generator employs a Variational Autoencoder (VAE)~\cite{kingma2013auto} trained on binary mask representations of 958 real archaeological fragments from the RePAIR dataset~\cite{tsesmelis2024re}. The architecture is visualized in Figure~\ref{fig:vae_architecture}.

\subsubsection{Encoder Architecture}

The encoder compresses 128×128 binary fragment masks into a 64-dimensional latent representation:

\begin{itemize}
    \item \textbf{Conv1}: $128 \times 128 \times 1 \rightarrow 64 \times 64 \times 32$ 
        \begin{itemize}
            \item 3×3 kernel, stride 2, padding 1
            \item ReLU activation, BatchNorm, Dropout(0.3)
        \end{itemize}
    \item \textbf{Conv2}: $64 \times 64 \times 32 \rightarrow 32 \times 32 \times 64$
        \begin{itemize}
            \item 3×3 kernel, stride 2, padding 1
            \item ReLU activation, BatchNorm, Dropout(0.3)
        \end{itemize}
    \item \textbf{Conv3}: $32 \times 32 \times 64 \rightarrow 16 \times 16 \times 128$
        \begin{itemize}
            \item 3×3 kernel, stride 2, padding 1
            \item ReLU activation, BatchNorm, Dropout(0.3)
        \end{itemize}
    \item \textbf{Conv4}: $16 \times 16 \times 128 \rightarrow 8 \times 8 \times 256$
        \begin{itemize}
            \item 3×3 kernel, stride 2, padding 1
            \item ReLU activation, BatchNorm, Dropout(0.3)
        \end{itemize}
    \item \textbf{Latent projection}: Flatten to 16,384-D, then project to 64-D $\mu$ and 64-D $\log\sigma^2$
\end{itemize}

\subsubsection{Decoder Architecture}

The decoder reconstructs fragment masks from 64-dimensional latent codes:

\begin{itemize}
    \item \textbf{Latent expansion}: 64-D $\rightarrow$ $8 \times 8 \times 256$ (reshape)
    \item \textbf{TransConv1}: $8 \times 8 \times 256 \rightarrow 16 \times 16 \times 128$
        \begin{itemize}
            \item 3×3 kernel, stride 2, padding 1, output\_padding 1
            \item ReLU activation, BatchNorm
        \end{itemize}
    \item \textbf{TransConv2}: $16 \times 16 \times 128 \rightarrow 32 \times 32 \times 64$
        \begin{itemize}
            \item 3×3 kernel, stride 2, padding 1, output\_padding 1
            \item ReLU activation, BatchNorm
        \end{itemize}
    \item \textbf{TransConv3}: $32 \times 32 \times 64 \rightarrow 64 \times 64 \times 32$
        \begin{itemize}
            \item 3×3 kernel, stride 2, padding 1, output\_padding 1
            \item ReLU activation, BatchNorm
        \end{itemize}
    \item \textbf{TransConv4}: $64 \times 64 \times 32 \rightarrow 128 \times 128 \times 1$
        \begin{itemize}
            \item 3×3 kernel, stride 2, padding 1, output\_padding 1
            \item Sigmoid activation (output in [0,1])
        \end{itemize}
\end{itemize}

\subsubsection{Training Configuration}

\begin{itemize}
    \item \textbf{Loss function}: $\mathcal{L} = \text{BCE}(x, \hat{x}) + \beta \cdot D_{\text{KL}}(q(z|x) \| \mathcal{N}(0, I))$ where $\beta=1.0$
    \item \textbf{Optimizer}: Adam with $\beta_1=0.9$, $\beta_2=0.999$, $\epsilon=10^{-8}$
    \item \textbf{Learning rate}: $10^{-4}$ (constant, no scheduling)
    \item \textbf{Batch size}: 32
    \item \textbf{Epochs}: 44 (early stopping based on validation loss)
    \item \textbf{Best validation loss}: 1623.76 (achieved at epoch 44)
    \item \textbf{Training data}: 958 binary masks from RePAIR dataset (80/10/10 train/val/test split)
    \item \textbf{Hardware}: NVIDIA RTX 4070 GPU (8GB VRAM).
\end{itemize}

%%%%%%%%%%%%%%%%%%%%%%%%%%%%%%%%%%%%%%%%%%%%%%%%%%%%%%%%%%%%%%%%%%%%%%%%%%%%%%%%
\subsection{Source Image Collection}
\label{sec:supp_met_collection}
%%%%%%%%%%%%%%%%%%%%%%%%%%%%%%%%%%%%%%%%%%%%%%%%%%%%%%%%%%%%%%%%%%%%%%%%%%%%%%%%

We utilize artwork images from The Metropolitan Museum of Art's Open Access collection~\cite{met_open_access}, accessed via their public API. The collection process ensures high-quality, diverse cultural heritage imagery suitable for synthetic archaeological puzzle generation.

\subsubsection{Collection Pipeline}

\begin{enumerate}
    \item \textbf{API Query}: Query \texttt{collectionapi.metmuseum.org} for public domain objects
    \item \textbf{Filtering}: Apply \texttt{isPublicDomain=True} AND \texttt{title NOT LIKE '\%fragment\%'}, in order to assure selected images are indeed categorized as public domain, while filtering out images of already fragmented artifacts. 
    \item \textbf{Sampling}: Random selection of 40,000 unique object IDs (20,000 for GAP-3, 20,000 for GAP-5)
    \item \textbf{Download}: Parallel retrieval with 20 workers and retry logic for failed requests
    \item \textbf{Storage}: Full-resolution primary images with complete metadata
    \item \textbf{Metadata}: CSV files with object ID, title, artist information, date/period, department, culture, medium, and dimensions (where available in the MET's original metadata)
\end{enumerate}

\subsubsection{Collection Diversity Statistics}

Analysis of the 40,000 collected images reveals exceptional temporal, geographical, and medium diversity:

\paragraph{Departmental Distribution:}
\begin{itemize}
    \item 19 unique departments represented
    \item Top 5: Drawings \& Prints (29.8\%), European Sculpture \& Decorative Arts (14.4\%), Asian Art (13.8\%), Greek \& Roman Art (5.7\%), Egyptian Art (5.4\%)
\end{itemize}

\paragraph{Temporal Coverage:}
\begin{itemize}
    \item Range: 970 BCE to 2000 CE ($\sim$2,970 years)
    \item Distribution: 19th century (24.6\%), 16th-17th centuries (11.7\%), 18th century (11.5\%), ancient-medieval periods (3.8\%), before 0 CE (2.6\%)
\end{itemize}

\paragraph{Media Representation:}
\begin{itemize}
    \item Prints (21.6\%), metalwork (15.4\%), textiles (10.4\%), ceramics (10.2\%), drawings (8.3\%), photographs (5.6\%), sculptures (3.6\%), paintings (2.1\%)
\end{itemize}

\paragraph{Cultural Origins:}
\begin{itemize}
    \item 1,933 unique cultures represented
    \item Top 5: Japanese (12.9\%), American (10.2\%), Chinese (8.1\%), French (7.0\%), Italian (3.0\%)
\end{itemize}

\noindent\textbf{Dataset Separation:} GAP-3 and GAP-5 use completely disjoint sets of 20,000 images each, ensuring independent evaluation without image overlap.

%%%%%%%%%%%%%%%%%%%%%%%%%%%%%%%%%%%%%%%%%%%%%%%%%%%%%%%%%%%%%%%%%%%%%%%%%%%%%%%%
\subsection{Statistical Validation}
\label{sec:supp_validation}
%%%%%%%%%%%%%%%%%%%%%%%%%%%%%%%%%%%%%%%%%%%%%%%%%%%%%%%%%%%%%%%%%%%%%%%%%%%%%%%%

We validate that VAE-generated fragments preserve the statistical distribution of real archaeological fragments through comprehensive shape analysis.

\subsubsection{Geometric Feature Definitions}

We formally define the eight geometric features extracted from fragment binary masks $M \in \{0,1\}^{H \times W}$:

\begin{enumerate}
    \item \textbf{Area} $A = \sum_{i,j} M(i,j)$: Total number of foreground pixels, providing an absolute size measure in px$^2$.
    
    \item \textbf{Perimeter} $P$: Length of the fragment boundary computed via contour tracing, measured in pixels. Captures edge extent and complexity.
    
    \item \textbf{Aspect Ratio} $r = w_{\text{bbox}}/h_{\text{bbox}}$: Ratio of minimum bounding rectangle width to height. Note that original fragments were normalized to square bounding boxes (aspect ratio $\approx 1$) pre-training to ensure consistent input dimensions (128×128 pixels), resulting in distributions centered near unity.
    
    \item \textbf{Solidity} $S = A/A_{\text{hull}}$: Ratio of fragment area to its convex hull area. $S = 1$ for convex fragments; $S < 1$ quantifies boundary concavity depth. Formally, $A_{\text{hull}} = \text{Area}(\text{ConvexHull}(M))$.
    
    \item \textbf{Circularity} $C = 4\pi A/P^2$: Isoperimetric quotient comparing shape to a circle. $C = 1$ for perfect circles; $C < 1$ for irregular shapes. Invariant to scaling.
    
    \item \textbf{Compactness} $K = P^2/A$: Inverse measure of shape efficiency. Lower values indicate more compact shapes; higher values reflect irregular boundaries. Related to circularity by $K = 4\pi/C$.
    
    \item \textbf{Vertices} $V$: Number of vertices in the convex hull approximation, computed via Douglas-Peucker algorithm with tolerance $\epsilon = 0.01P$. Represents corner count and polygon complexity.
    
    \item \textbf{Concavities} $N$: Number of contour points exhibiting negative curvature (inward bending), computed via discrete derivative approximation: $N = |\{p \in \partial M : \kappa(p) < -\tau\}|$ where $\kappa(p)$ is local curvature and $\tau$ is a small threshold. Quantifies edge irregularity.
\end{enumerate}

\noindent These features capture complementary aspects of fragment morphology: global size (area), boundary characteristics (perimeter, circularity, compactness), shape regularity (solidity, aspect ratio), and fine-scale structure (vertices, concavities).

\subsubsection{Summary Statistics}

Table~\ref{tab:vae_statistics_detailed} presents comprehensive statistics comparing real and synthetic fragments across all features.

\begin{table*}[t]
\centering
\caption{Detailed summary statistics comparing real (RePAIR) and synthetic (VAE-generated) fragments across eight geometric features (N=958 each).}
\label{tab:vae_statistics_detailed}
\small
\begin{tabular}{lccccc}
\toprule
\textbf{Feature} & \textbf{Group} & \textbf{Mean $\pm$ SD} & \textbf{Median} & \textbf{Min--Max} & \textbf{IQR} \\
\midrule
\multirow{2}{*}{Area (px$^2$)} 
  & Real & $10{,}617 \pm 1{,}333$ & 10,598 & 7,245--14,821 & 1,746 \\
  & Synthetic & $10{,}716 \pm 863$ & 10,705 & 8,012--13,254 & 1,142 \\
\midrule
\multirow{2}{*}{Perimeter (px)} 
  & Real & $451.95 \pm 29.72$ & 449.80 & 370.2--542.6 & 39.8 \\
  & Synthetic & $434.88 \pm 27.60$ & 433.10 & 361.7--518.4 & 36.2 \\
\midrule
\multirow{2}{*}{Aspect Ratio} 
  & Real & $1.001 \pm 0.023$ & 0.998 & 0.945--1.089 & 0.031 \\
  & Synthetic & $0.996 \pm 0.055$ & 0.993 & 0.842--1.178 & 0.072 \\
\midrule
\multirow{2}{*}{Solidity} 
  & Real & $0.931 \pm 0.027$ & 0.935 & 0.845--0.986 & 0.036 \\
  & Synthetic & $0.936 \pm 0.030$ & 0.941 & 0.831--0.991 & 0.040 \\
\midrule
\multirow{2}{*}{Circularity} 
  & Real & $0.655 \pm 0.068$ & 0.661 & 0.471--0.812 & 0.089 \\
  & Synthetic & $0.718 \pm 0.081$ & 0.726 & 0.504--0.895 & 0.107 \\
\midrule
\multirow{2}{*}{Compactness} 
  & Real & $19.47 \pm 2.90$ & 19.23 & 13.12--28.61 & 3.81 \\
  & Synthetic & $17.78 \pm 2.41$ & 17.53 & 11.92--25.42 & 3.16 \\
\midrule
\multirow{2}{*}{Vertices} 
  & Real & $9.73 \pm 2.12$ & 10 & 5--16 & 3 \\
  & Synthetic & $11.78 \pm 2.03$ & 12 & 6--18 & 3 \\
\midrule
\multirow{2}{*}{Concavities} 
  & Real & $19.79 \pm 3.62$ & 20 & 11--31 & 5 \\
  & Synthetic & $23.01 \pm 3.74$ & 23 & 13--35 & 5 \\
\bottomrule
\end{tabular}
\end{table*}

\subsubsection{Box Plot Visualizations}

\begin{figure*}[t]
    \centering
    \includegraphics[width=\textwidth]{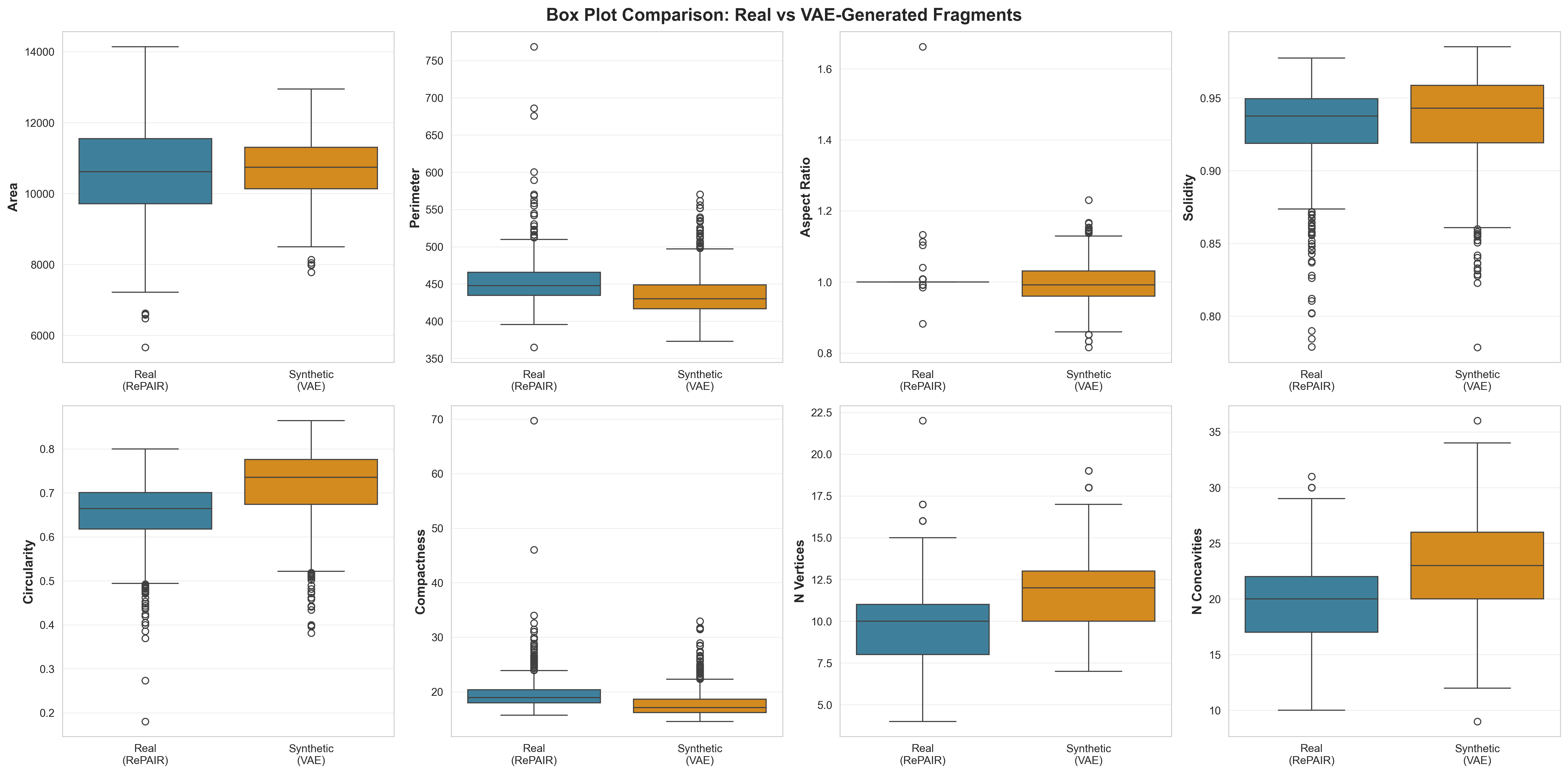}
    \caption{\textbf{Distribution comparison via box plots.} Real (RePAIR) fragments shown in blue, synthetic (VAE) fragments in orange. Boxes indicate interquartile ranges (IQR), horizontal lines show medians, whiskers extend to 1.5×IQR, and circles represent outliers. Core shape properties (area, solidity) exhibit high similarity, while edge complexity metrics show expected smoothing effects from VAE reconstruction.}
    \label{fig:vae_boxplots_supp}
\end{figure*}

Figure~\ref{fig:vae_boxplots_supp} presents box plots showing medians, interquartile ranges, and outliers for all features, demonstrating the close alignment between real and synthetic fragment distributions.

\subsubsection{Dimensionality Reduction Analysis}

Principal Component Analysis (PCA) on the 8-dimensional feature space reveals:

\begin{itemize}
    \item \textbf{PC1 (45.4\% variance)}: Overall fragment size and complexity (high loadings: area, perimeter, vertices, concavities)
    \item \textbf{PC2 (17.8\% variance)}: Edge irregularity and compactness (high loadings: circularity, compactness)
    \item \textbf{PC3 (13.1\% variance)}: Aspect ratio and orientation
    \item \textbf{PC4--PC8 (23.7\% variance)}: Higher-order shape variations
\end{itemize}

The first two principal components capture 63.2\% of total variance. As shown in Figure~\ref{fig:vae_embedding} (main paper), real and synthetic fragments exhibit substantial overlap in this reduced space, with no isolated clusters or systematic biases, confirming that the VAE successfully captures the underlying distribution of archaeological fragment shapes.

These validation results confirm that our VAE successfully captures the geometric essence of archaeological fragments while maintaining practical advantages for large-scale dataset generation. High fidelity in core shape features (1-3\% differences in area and solidity) and substantial PCA overlap demonstrate that GAP fragments authentically represent real artifact morphology. Expected smoothing in edge complexity metrics reflects VAE reconstruction characteristics but preserves the irregular, non-linear erosion patterns absent from existing square-piece datasets. This combination of archaeological realism and synthetic scalability positions GAP as an effective bridge between simplified academic benchmarks and real-world heritage reconstruction, enabling systematic algorithm development on challenging, realistic fragment geometries at a scale impossible with limited authentic artifact collections.

%%%%%%%%%%%%%%%%%%%%%%%%%%%%%%%%%%%%%%%%%%%%%%%%%%%%%%%%%%%%%%%%%%%%%%%%%%%%%%%%
\section{Implementation Details: Models and Baselines}
\label{sec:implementation_supplementary}
%%%%%%%%%%%%%%%%%%%%%%%%%%%%%%%%%%%%%%%%%%%%%%%%%%%%%%%%%%%%%%%%%%%%%%%%%%%%%%%%

This section provides complete implementation details for PuzzleFlow and all baseline methods, enabling full reproducibility of our experimental results.

%%%%%%%%%%%%%%%%%%%%%%%%%%%%%%%%%%%%%%%%%%%%%%%%%%%%%%%%%%%%%%%%%%%%%%%%%%%%%%%%
\subsection{PuzzleFlow: Architecture and Training}
\label{sec:supp_puzzleflow}
%%%%%%%%%%%%%%%%%%%%%%%%%%%%%%%%%%%%%%%%%%%%%%%%%%%%%%%%%%%%%%%%%%%%%%%%%%%%%%%%

\begin{figure*}[t]
    \centering
    \includegraphics[width=\linewidth]{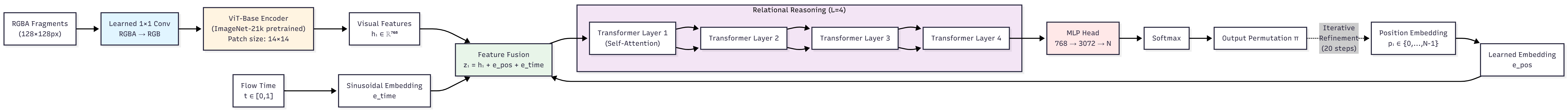}
    \caption{\textbf{PuzzleFlow Architecture.} Individual puzzle fragments are processed through a pretrained ViT backbone to extract 768-dimensional visual features. These features are combined with position embeddings (encoding current fragment placements) and time embeddings (encoding flow matching timestep), then passed through 4 additional transformer layers for cross-piece reasoning. The output head predicts logits over all possible positions for each fragment. During training, we sample random timesteps $t \in [0,1]$ and interpolate between scrambled and solved states. During inference, we iteratively denoise from random initialization to the solved configuration.}
    \label{fig:puzzleflow_architecture}
\end{figure*}

PuzzleFlow combines a pretrained Vision Transformer backbone with additional transformer layers and discrete flow matching for puzzle reassembly. The architecture is visualized in Figure~\ref{fig:puzzleflow_architecture}.

\subsubsection{Architecture Specifications}

Table~\ref{tab:architecture_details} details the complete PuzzleFlow architecture.

\begin{table}[h]
\centering
\caption{PuzzleFlow architecture specifications. Configuration is identical across GAP-3 and GAP-5 datasets.}
\label{tab:architecture_details}
\small
\begin{tabular}{ll}
\toprule
\textbf{Component} & \textbf{Configuration} \\
\midrule
\multicolumn{2}{l}{\textit{Visual Encoder}} \\
\quad Backbone & ViT-Base-Patch16-224~\cite{dosovitskiy2020image} \\
\quad Initialization & Pretrained on ImageNet-21k \\
\quad Hidden dimension & 768 \\
\quad Attention heads (ViT) & 12 \\
\quad Encoder layers (ViT) & 12 \\
\quad Patch size & 16×16 pixels \\
\quad Input resolution & 224×224 per fragment \\
\quad Gradient checkpointing & Enabled \\
\midrule
\multicolumn{2}{l}{\textit{Transformer Decoder}} \\
\quad Hidden dimension ($d_\text{model}$) & 768 \\
\quad Attention heads & 12 \\
\quad Decoder layers & 4 \\
\quad Feedforward dimension & 3072 (4×$d_\text{model}$) \\
\quad Dropout rate & 0.1 \\
\midrule
\multicolumn{2}{l}{\textit{Flow Matching}} \\
\quad Interpolation type & Linear \\
\quad Time embedding & Sinusoidal (dim=768) \\
\quad Position embedding & 2-layer MLP (2 → 256 → 128) \\
\quad Output head & 3-layer MLP (768 → 512 → 256 → $N$) \\
\midrule
\multicolumn{2}{l}{\textit{Dataset-Specific Parameters}} \\
\quad GAP-3 fragment size & 128×128 pixels \\
\quad GAP-5 fragment size & 128×128 pixels \\
\quad Solved image size (GAP-3) & 384×384 pixels (3×128) \\
\quad Solved image size (GAP-5) & 640×640 pixels (5×128) \\
\quad Grid size (GAP-3) & 3×3 (9 positions) \\
\quad Grid size (GAP-5) & 5×5 (25 positions) \\
\bottomrule
\end{tabular}
\end{table}

\subsubsection{Training Hyperparameters}

All training was conducted on NVIDIA RTX 4090 GPUs with 24GB memory. Table~\ref{tab:training_details} provides complete training configuration.

\begin{table}[h]
\centering
\caption{PuzzleFlow training hyperparameters. Configuration is consistent across both GAP-3 and GAP-5 datasets.}
\label{tab:training_details}
\small
\begin{tabular}{ll}
\toprule
\textbf{Parameter} & \textbf{Value} \\
\midrule
\multicolumn{2}{l}{\textit{Optimization}} \\
\quad Optimizer & AdamW~\cite{loshchilov2017decoupled} \\
\quad Learning rate & $1 \times 10^{-5}$ \\
\quad Weight decay & 0.01 \\
\quad Betas & (0.9, 0.999) \\
\quad Learning rate schedule & OneCycleLR \\
\quad Warmup percentage & 10\% \\
\quad Gradient clipping & None \\
\midrule
\multicolumn{2}{l}{\textit{Training Configuration}} \\
\quad Number of epochs & 30 \\
\quad Batch size (training) & 8 \\
\quad Batch size (validation) & 2 \\
\quad Mixed precision (AMP) & Enabled (FP16) \\
\quad Validation flow steps & 5 (20 for final eval) \\
\midrule
\multicolumn{2}{l}{\textit{Data Loading}} \\
\quad Number of workers & 8 \\
\quad Persistent workers & Enabled \\
\quad Prefetch factor & 2 \\
\quad Pin memory & Enabled \\
\midrule
\multicolumn{2}{l}{\textit{Regularization}} \\
\quad ViT backbone & Fine-tuned (unfrozen) \\
\quad Frozen ViT layers & 0 \\
\quad Dropout & 0.1 \\
\bottomrule
\end{tabular}
\end{table}

\subsubsection{Loss Function and Training Objective}

The training objective combines cross-entropy loss over predicted positions at randomly sampled timesteps $t \in [0, 1]$ during the flow process:
\begin{equation}
\mathcal{L} = \mathbb{E}_{t \sim \mathcal{U}(0,1), \mathbf{x}_0, \mathbf{x}_1} \left[ -\sum_{i=1}^{N} \log p_\theta(x_1^{(i)} | \mathbf{x}_t, t) \right]
\end{equation}
where $\mathbf{x}_0$ represents the initial scrambled permutation, $\mathbf{x}_1$ is the target solved configuration, $\mathbf{x}_t = (1-t)\mathbf{x}_0 + t\mathbf{x}_1$ is the linearly interpolated state, and $N$ is the number of fragments.

\subsubsection{Implementation Optimizations}

Several techniques were utilized for efficient training:

\begin{itemize}
    \item \textbf{Gradient checkpointing}: Reduces memory usage by $\sim$30\% by recomputing activations during the backward pass rather than storing them.
    \item \textbf{Mixed precision training}: Automatic Mixed Precision (AMP) with FP16 enabled, providing 1.5--2× speedup and 40\% memory reduction while maintaining numerical stability through automatic loss scaling.
    \item \textbf{Adaptive batch sizing}: Training uses batch size 8, but validation uses batch size 2 to prevent out-of-memory errors during the multi-step sampling process.
    \item \textbf{Fast validation}: During training, validation uses 5 flow steps for speed; final evaluation uses 20 steps.
\end{itemize}

\subsection{Learning-Based Baseline Adaptations}
\label{sec:supp_learning_baselines}

We evaluate three state-of-the-art learning-based methods: FCViT~\cite{kim2025solving}, JPDVT~\cite{liu2024solving}, and PuzLM~\cite{elkin2025seq}. Since FCViT and JPDVT were originally designed for square RGB puzzles with internal shuffling mechanisms, we reconstructed solved puzzle images by placing GAP's RGBA fragments (with alpha channel dropped) at their ground truth grid positions, resized to method-specific dimensions. PuzLM operates on individual scrambled fragments and required only conversion from RGBA to RGB format. All methods were trained for 30 epochs using default hyperparameters from their official repositories. For JPDVT, we used JPDVT-T variant.

\paragraph{Note on Preprocessing Rationale.}
This preprocessing step is necessary because GAP fragments are provided as individual RGBA images with irregular shapes, whereas the baseline methods expect complete grid-aligned images that they internally shuffle during training. By reconstructing solved puzzles and allowing each method to perform its own internal shuffling, we ensure fair comparison under each method's original design assumptions.

\subsection{Classical Baseline Implementations}
\label{sec:supp_classical}
%%%%%%%%%%%%%%%%%%%%%%%%%%%%%%%%%%%%%%%%%%%%%%%%%%%%%%%%%%%%%%%%%%%%%%%%%%%%%%%%

\subsubsection{Greedy Solver (Pomeranz et al. 2011)}
\label{sec:supp_greedy}

We implemented the fully automated greedy solver of Pomeranz~\textit{et al.}~\cite{pomeranz2011fully}, which constructs puzzles through iterative best-buddy placement.

\paragraph{Compatibility Metric.}
We compute pairwise dissimilarity in LAB color space. For adjacent pieces $x_i$ and $x_j$, the dissimilarity along edge direction $r$ is:
\begin{equation}
D(x_i, x_j, r) = \sum_k \left( |2e_i^k - e_i^{k-1} - e_j^k|^P + |2e_j^k - e_j^{k+1} - e_i^k|^P \right)^{Q/P}
\end{equation}
where $e_i^k$ denotes the $k$-th pixel along the edge of piece $x_i$, $r \in \{\text{LEFT, RIGHT, UP, DOWN}\}$, and $P=0.3$, $Q=0.0625$ are constants from~\cite{pomeranz2011fully}.

Compatibility is computed as:
\begin{equation}
C(x_i, x_j, r) = \exp\left(-\frac{D(x_i, x_j, r)}{\text{percentile}_{25}(D(x_i, \cdot, r))}\right)
\end{equation}

\paragraph{Best Buddy Definition.}
Pieces $x_i$ and $x_j$ are \textit{best buddies} in direction $r$ if:
\begin{equation}
\argmax_{k} C(x_i, k, r) = j \quad \text{and} \quad \argmax_{k} C(x_j, k, \bar{r}) = i
\end{equation}
where $\bar{r}$ denotes the opposite direction.

\paragraph{Algorithm.}
The solver proceeds in three phases:
\begin{enumerate}
    \item \textbf{Seed selection}: Choose the piece with maximum mutual best buddies as initial seed, placed at grid center.
    \item \textbf{Greedy placement}: Iteratively select candidate slots (positions adjacent to placed pieces with maximum occupied neighbors) and assign pieces with highest average compatibility. Ties are broken using best-buddy relationships.
    \item \textbf{Refinement}: Segment assembly into connected components based on best-buddy relationships. Keep only the largest segment, re-center, and repeat until no improvement in the best-buddies metric (BBM).
\end{enumerate}

The best-buddies metric evaluates solution quality:
\begin{equation}
\text{BBM} = \frac{\text{\# best-buddy edges}}{\text{\# total edges}}
\end{equation}

\paragraph{Implementation Details.}
\begin{itemize}
    \item Input: Fragment images converted to LAB color space using \texttt{scikit-image}
    \item Precomputation: Full $4 \times N \times N$ dissimilarity matrix for all directions
    \item Grid handling: Dynamic expansion via NumPy array rolling when boundaries are reached
    \item Termination: Algorithm stops when BBM no longer improves
\end{itemize}

\paragraph{Computational Complexity.}
Building the dissimilarity matrix requires $O(N^2 \cdot H)$ operations for $N$ pieces of height $H$ pixels. The placement phase is $O(N \cdot k)$ where $k$ is the number of iterations (typically 3--5). Runtime: 1--2 minutes for 3×3 puzzles, 5--10 minutes for 5×5 puzzles on a single CPU core.

\subsubsection{Genetic Algorithm Solver (Sholomon et al. 2013)}
\label{sec:supp_genetic}

We implemented the genetic algorithm approach of Sholomon~\textit{et al.}~\cite{sholomon2013genetic}, which frames puzzle solving as permutation optimization.

\paragraph{Representation.}
Each individual is a permutation $\pi \in S_N$ of piece indices, where position $i$ contains piece $\pi(i)$.

\paragraph{Fitness Function.}
Fitness is the negative sum of dissimilarities between adjacent pieces:
\begin{equation}
f(\pi) = -\sum_{(i,j) \text{ adjacent}} D(\pi(i), \pi(j), r_{i \to j})
\end{equation}
Higher fitness (lower total dissimilarity) indicates better solutions.

\paragraph{Genetic Operators.}
\begin{itemize}
    \item \textbf{Selection}: Tournament selection with tournament size 3
    \item \textbf{Crossover}: Partially Mapped Crossover (PMX) with rate 0.8
    \item \textbf{Mutation}: Three types (swap, inversion, scramble) with rate 0.01
    \item \textbf{Elitism}: Retain top 10\% unchanged
\end{itemize}

\paragraph{Algorithm Parameters.}
\begin{itemize}
    \item Population size: 100
    \item Maximum generations: 1000
    \item Early stopping: 100 generations without improvement
    \item Mutation rate: 0.01
    \item Crossover rate: 0.8
    \item Elitism ratio: 0.1
\end{itemize}

\paragraph{Computational Complexity.}
Each generation requires $O(P \cdot G^2)$ fitness evaluations for population size $P$ and grid size $G$. Total complexity is $O(T \cdot P \cdot G^2)$ for $T$ generations. Runtime: 2--5 minutes for 3×3 puzzles, 15--30 minutes for 5×5 puzzles on a single CPU core.

\subsubsection{Adaptation to GAP Datasets}

Both classical methods were adapted to handle GAP's irregular fragments:
\begin{itemize}
    \item \textbf{Color space conversion}: RGBA → RGB → LAB using \texttt{scikit-image}
    \item \textbf{Edge handling}: Dissimilarity computed along detected fragment boundaries (non-zero alpha channel regions)
    \item \textbf{Erosion robustness}: No special handling for erosion; methods rely purely on boundary compatibility, which degrades as erosion increases
\end{itemize}

The primary limitation of these classical approaches on GAP is their reliance on edge continuity. As erosion removes original boundaries, the compatibility metrics become less informative, leading to degraded performance compared to learning-based methods that leverage global visual patterns.

%%%%%%%%%%%%%%%%%%%%%%%%%%%%%%%%%%%%%%%%%%%%%%%%%%%%%%%%%%%%%%%%%%%%%%%%%%%%%%%%
\subsection{Evaluation Protocol}
\label{sec:supp_evaluation}
%%%%%%%%%%%%%%%%%%%%%%%%%%%%%%%%%%%%%%%%%%%%%%%%%%%%%%%%%%%%%%%%%%%%%%%%%%%%%%%%

All methods are evaluated using consistent metrics on held-out test sets:

\begin{itemize}
    \item \textbf{Exact Match Rate (Perfect Accuracy)}: Percentage of puzzles with all pieces correctly placed.
    \item \textbf{Position Accuracy (Direct Accuracy)}: Average fraction of correctly placed pieces per puzzle.
    \item \textbf{Spatial Relationship Accuracy (SRA)}: Average fraction of correctly adjacent piece pairs, as defined in the main paper.
\end{itemize}

For PuzzleFlow and JPDVT, we use 20-step sampling during evaluation. Classical methods produce deterministic outputs.

\section{Validation of PuzzleFlow on simpler settings}
\label{sup:jpwleg_validation}
Although PuzzleFlow is introduced mainly as a strong reference solver rather than the core focus of this work, have trained and evaluated this framework on the widely used JPwLEG-3 and JPwLEG-5 datasets, without any dataset-specific tuning, achieving absolute accuracy (AA) of 0.726 and perfect accuracy (PA) of 0.437 on JPwLEG-3; and AA of 0.290 and PA of 0 on JPwLEG-5. While these results do not outperform the SOTA on this benchmark, they are competitive with, and in some cases surpass, recent prominent approaches such as JPDVT, despite being obtained by a solver \textit{not specialized} to square-piece setting. 

Importantly, the performance is consistent with that observed on our corresponding GAP datasets, supporting that the proposed permutation-flow formulation is robust across both irregular and conventional square-piece puzzles rather than overfitting to GAP alone. See Table.~\ref{tab:res-jpwleg} for full results.

\begin{table}[t]
\centering
\caption{Validation on square-piece JPwLEG benchmarks. Some approaches did not report performance in all metrics, or in both dataset variations}
\label{tab:res-jpwleg}
\begin{small}
\begin{tabular*}{\textwidth}{@{\extracolsep{\fill}}lcccc}
\toprule
\multirow{2}{*}{\textbf{Method}} & \multicolumn{2}{c}{\textbf{JPwLEG-3} ($3\times3$)} & \multicolumn{2}{c}{\textbf{JPwLEG-5} ($5\times5$)} \\
\cmidrule(lr){2-3} \cmidrule(lr){4-5}
& \textbf{AA (\%)} & \textbf{PA (\%)} & \textbf{AA (\%)} & \textbf{PA (\%)} \\
\midrule
JPDVT~\cite{liu2024solving} & 71.3 & N/A & N/A & N/A \\
Greedy~\cite{pomeranz2011fully} & 71.5 & 40.0 & 24.1 & 0.1 \\
Deepzzle~\cite{paumard2020deepzzle} & 74.0 & 44.9 & 21.9 & 0.0 \\
Tabu~\cite{adamczewski2015discrete} & 73.8 & 44.8 & 24.6 & 0.0 \\
GA~\cite{sholomon2013genetic} & 73.9 & 44.9 & 25.1 & 0.0 \\
\textbf{PuzzleFlow (ours)} & 72.6 & 43.7 & 29.0 & 0.0 \\
SD$^2$RL~\cite{song2023siamese} & 81.6 & 59.7 & 40.3 & 5.1 \\
FCViT~\cite{kim2025solving} & \textbf{96.9} & \textbf{87.9} & N/A & N/A \\
PDN-GA~\cite{song2023solving} & 81.3 & 58.2 & 44.3 & 6.1 \\
ERL-MPP~\cite{song2025erlmpp} & N/A & N/A & 52.7 & 18.6 \\
VLHSA~\cite{xu2025vlhsa} & 85.4 & N/A & \underline{66.9} & \underline{19.0} \\
PuzLM~\cite{elkin2025seq} & \underline{91.9} & \underline{84.5} & \textbf{72.1} & \textbf{32.5} \\
\bottomrule
\end{tabular*}
\end{small}
\end{table}

\section{Qualitative Results}
\label{sec:qualitative_results}
%%%%%%%%%%%%%%%%%%%%%%%%%%%%%%%%%%%%%%%%%%%%%%%%%%%%%%%%%%%%%%%%%%%%%%%%%%%%%%%%

\begin{figure*}[t]
    \centering
    \includegraphics[width=\textwidth]{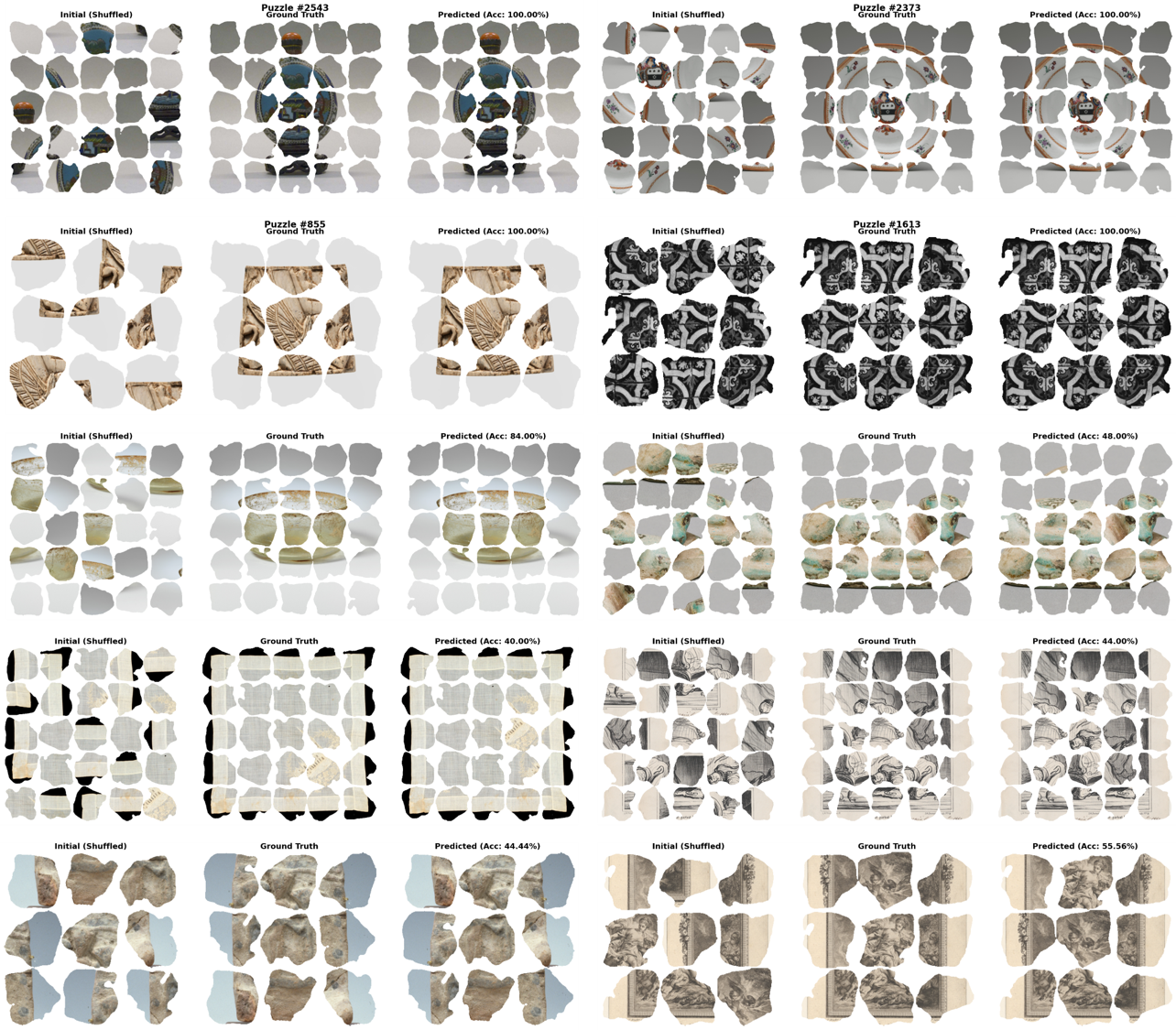}
    \caption{\textbf{Qualitative Results.} Representative examples of PuzzleFlow solving GAP puzzles. \textbf{Top rows:} Successful reconstructions on GAP-3 (left) and GAP-5 (right) with heavily eroded fragments. \textbf{Bottom rows:} Challenging failure cases where erosion or visual ambiguity leads to errors. PuzzleFlow successfully handles irregular fragment geometries and leverages global visual patterns, though some puzzles with extreme erosion or repetitive textures remain challenging.}
    \label{fig:qualitative_results}
\end{figure*}

Figure~\ref{fig:qualitative_results} shows representative examples of PuzzleFlow solving GAP-3 and GAP-5 puzzles, including both successful reconstructions and challenging failure cases. Note that in some cases generally low evaluation scores could be explained by relatively similar pieces, mistakenly assigned to the correspondent position.  

\subsection{Ethical Statement} 
\label{sup:ethical_statement}
Our research uses only publicly available artifact images and metadata from the MET collection (CC0 license), ensuring compliance with data ownership and privacy regulations. The fragments in our newly presented GAP dataset are synthetic, generated without any human-derived personal information or sensitive content. We believe the potential for misuse is minimal; however, we acknowledge that automatic assembly tools could theoretically be misapplied to heritage items without proper authority. We encourage responsible use strictly within permitted conservation, restoration, and academic boundaries. All datasets and code will be released according to museum guidelines and community standards.

\subsection{Code and Data Availability}

Complete implementation code and the GAP datasets have been publically released. The codebase includes: The complete GAP datasets, along with the generation code (VAE training, fragment synthesis) and trained VAE checkpoint, PuzzleFlow training and inference scripts, and some adapted baseline implementations.

% \begin{itemize}
%     \item PuzzleFlow training and inference scripts
%     \item Adapted baseline implementations
%     \item The complete GAP datasets, along with the generation code (VAE training, fragment synthesis) and trained VAE checkpoint.
%     \item Evaluation utilities
% \end{itemize}

\end{document}